\documentclass[]{fairmeta}

\usepackage{microtype}
\usepackage{graphicx}
\usepackage{subcaption}
\usepackage{csquotes}
\usepackage{afterpage}

\usepackage{amsmath}
\usepackage{amssymb}
\usepackage{mathtools}
\usepackage{amsthm}
\usepackage{xspace}
\usepackage{colortbl}
\usepackage{multirow}
\usepackage{enumitem}
\usepackage{wrapfig}
\usepackage{nicefrac}

\newcommand{\model}{V-JEPA~2\xspace}
\newcommand{\oldModel}{V-JEPA\xspace}
\newcommand{\acModel}{V-JEPA~2-AC\xspace}

\definecolor{fbApp}{HTML}{c8e7fa}
\definecolor{fbPurple3}{HTML}{f0ebf5}

\definecolor{citecolor}{HTML}{0071BC}
\definecolor{linkcolor}{HTML}{ED1C24}

\definecolor{citecolor}{HTML}{0071BC}
\definecolor{linkcolor}{HTML}{ED1C24}

\title{V-JEPA~2: Self-Supervised Video Models Enable Understanding, Prediction and Planning}

\author[1,*]{Mahmoud Assran}
\author[1,*]{Adrien Bardes}
\author[1,*]{David Fan}
\author[1,*]{Quentin Garrido}
\author[1,*]{Russell Howes}
\author[1,*]{Mojtaba Komeili}
\author[1,*]{Matthew Muckley}
\author[1,*]{Ammar Rizvi}
\author[1,*]{Claire Roberts}
\author[1,*]{Koustuv Sinha}
\author[1,2,*]{Artem Zholus}
\author[1,*]{Sergio Arnaud}
\author[1,*]{Abha Gejji}
\author[1,*]{Ada Martin}
\author[1,*]{Francois Robert Hogan}
\author[1,*]{Daniel Dugas}
\author[1]{Piotr Bojanowski}
\author[1]{Vasil Khalidov}
\author[1]{Patrick Labatut}
\author[1]{Francisco Massa}
\author[1]{Marc Szafraniec}
\author[1]{Kapil Krishnakumar}
\author[1]{Yong Li}
\author[1]{Xiaodong Ma}
\author[2]{Sarath Chandar}
\author[1,*]{Franziska Meier}
\author[1,*]{Yann LeCun}
\author[1,*]{Michael Rabbat}
\author[1,*]{Nicolas Ballas}

\affiliation[1]{FAIR at Meta}
\affiliation[2]{Mila -- Quebec AI Institute and Polytechnique Montr\'{e}al}

\contribution[*]{Core Team}

\abstract{
A major challenge for modern AI is to learn to understand the world and learn to act largely by observation~\citep{lecun2022path}.
This paper explores a self-supervised approach that combines internet-scale video data with a small amount of interaction data (robot trajectories), to develop models capable of understanding, predicting, and planning in the physical world. We first pre-train an action-free joint-embedding-predictive architecture, \model, on a video and image dataset comprising over 1 million hours of internet video. \model achieves strong performance on motion understanding (77.3 top-1 accuracy on Something-Something v2) and state-of-the-art performance on human action anticipation (39.7 recall-at-5 on Epic-Kitchens-100) surpassing previous task-specific models. Additionally, after aligning \model with a large language model, we demonstrate state-of-the-art performance on multiple video question-answering tasks at the 8 billion parameter scale (e.g., 84.0 on PerceptionTest, 76.9 on TempCompass). Finally, we show how self-supervised learning can be applied to robotic planning tasks by post-training a latent action-conditioned world model, \acModel, using less than 62 hours of unlabeled robot videos from the Droid dataset. We deploy \acModel zero-shot on Franka arms in two different labs and enable picking and placing of objects using planning with image goals. Notably, this is achieved without collecting any data from the robots in these environments, and without any task-specific training or reward. This work demonstrates how self-supervised learning from web-scale data and a small amount of robot interaction data can yield a world model capable of planning in the physical world.
}
  
\date{\today}
\correspondence{Nicolas Ballas <\email{ballasn@meta.com}> and Michael Rabbat <\email{mikerabbat@meta.com}>}

\metadata[Code]{\url{https://github.com/facebookresearch/vjepa2}}
\metadata[Blogpost]{\url{https://ai.meta.com/blog/v-jepa-2-world-model-benchmarks}}

\begin{document}

\maketitle

\section{Introduction}
\label{section:intro}

Humans have the ability to adapt and generalize when taking on new tasks and operating in unfamiliar environments.
Several cognitive learning theories suggest that humans learn an internal model of the world by integrating low-level sensory inputs to represent and predict future states~\citep{craik1967nature,rao1999predictive}, and they further posit that this world model shapes our perception at any given moment, playing a crucial role in informing our understanding of reality~\citep{friston2010free,clark2013whatever,nortmann2015primary}.
Moreover, our ability to predict the effects of our actions on future states of the world is also essential for goal-oriented planning~\citep{sutton1981adaptive,sutton1998reinforcement,ha2018world,wolpert2000computational}.
Building artificial agents that learn a world model from sensory data, such as video, could enable them to \emph{understand} the physical world, \emph{predict} future states, and effectively --- like humans --- \emph{plan} in new situations, resulting in systems capable of tackling tasks that have not been encountered before.

\begin{figure}[!t]
    \centering
    \includegraphics[width=\linewidth]{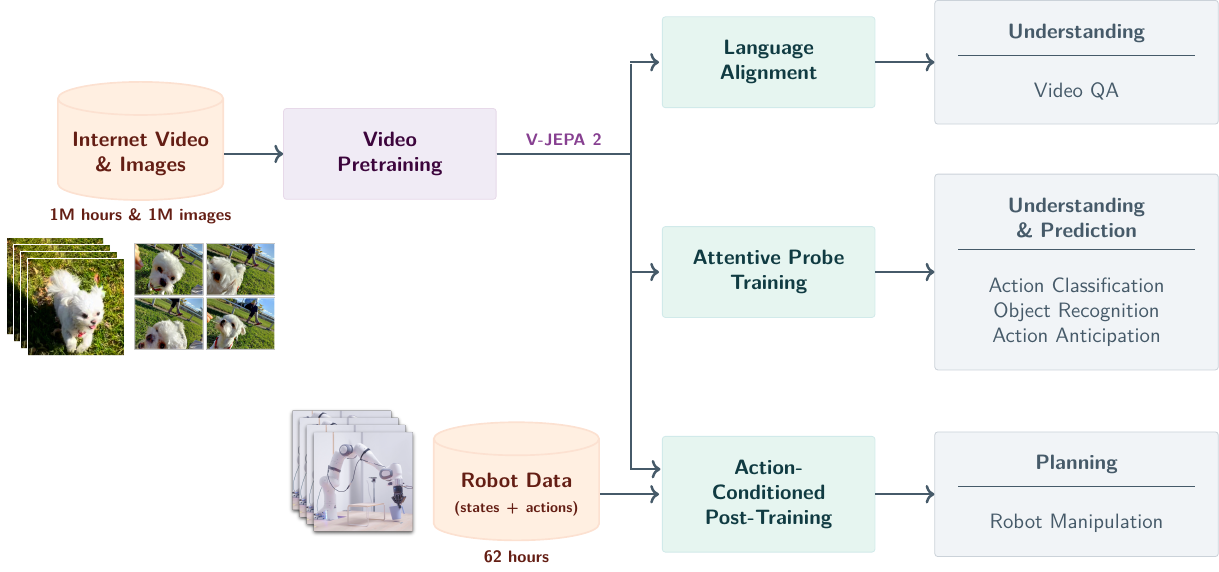}
    \caption{{\bf \model Overview.}
    Leveraging 1M hours of internet-scale video and 1M images, we pretrain the \model video model using a visual mask denoising objective~\citep{bardes2024revisiting, assran2023self}, and leverage this model for downstream tasks such as action classification, object recognition, action anticipation, and Video Question Answering by aligning the model with an LLM backbone.
    After pretraining, we can also freeze the video encoder and train a new action-conditioned predictor with a small amount of robot interaction data on top of the learned representations, and leverage this action-conditioned model, \acModel, for downstream robot manipulation tasks using planning within a model predictive control loop.}
    \label{fig:overview}
\end{figure}

Previous works have explored the development of predictive world models from interaction data consisting of state-action sequences, often also relying on explicit reward feedback from the environment to infer goals~\citep{sutton1981adaptive,fragkiadaki2015learning,ha2018world,hafner2019learning,hansen2022temporal}. However, the limited availability of real-world interaction data constrains the scalability of these methods.
To address this limitation, more recent works have leveraged both internet-scale video and interaction data towards training action-conditioned video generation models for robot control, but only demonstrate limited results in robot execution using model-based control~\citep{hu2023gaia,yang2024learning,bruce2024genie,agarwal2025cosmos}.
In particular, this line of research often emphasizes the evaluation of the faithfulness of the predictions and visual quality instead of planning capabilities, perhaps due to the computational cost of planning by generating video.

In this work, we build upon the self-supervised hypothesis as a means to learn world models that capture background knowledge of the world largely from observation. Specifically, we leverage the joint-embedding predictive architecture (JEPA)~\citep{lecun2022path}, which learns by making predictions in a learned representation space. 
In contrast to approaches that focus on learning entirely from interaction data, self-supervised learning enables us to make use of internet-scale video --- depicting sequences of states without direct observations of the actions ---
to learn to both represent video observations and learn a predictive model for world dynamics in this learned representation space.
Furthermore, in contrast to approaches based on video generation, the JEPA approach focuses on learning representations for predictable aspects of a scene (e.g., the trajectory of an object in motion) while ignoring unpredictable details that generative objectives emphasize, since they make pixel-level predictions (e.g., the precise location of each blade of grass in a field, or each leaf on a tree).
By scaling JEPA pretraining, we demonstrate that it yields video representations with state-of-the-art understanding and prediction capabilities, and that such representations can be leveraged as a basis for action-conditioned predictive models and enable zero-shot planning. 

Our approach, \model, utilizes a stage-wise training procedure, beginning with action-free pre-training on internet-scale video, followed by post-training with a small amount of interaction data (see \Cref{fig:overview}).
In the first stage, we use a mask-denoising feature prediction objective~\citep{assran2023self,bardes2024revisiting}, where the model predicts masked segments of a video in a learned representation space.
We train the \model  encoder with up to 1 billion parameters and with more than 1 million hours of video.
Our experiments confirm that scaling self-supervised  video pretraining enhances the encoder's ability to achieve visual understanding, including broad motion and appearance recognition capabilities, through probe-based evaluations and by aligning the encoder with a language model for video question-answering \citep{benno2024, patraucean2023perception, liu2024tempcompass, cai2024temporalbench, shangguan2024tomato}.

Following pretraining on internet-scale video, we train an action-conditioned world model, \acModel, on a small set of interaction data using the representations learned in the first stage.
Our action-conditioned world model is a 300M-parameter transformer network employing a block-causal attention mechanism, which autoregressively predicts the representation of the next video frame conditioned on an action and previous states.
With as little as 62 hours of unlabeled interaction data from the Droid dataset~\citep{khazatsky2024droid}, we demonstrate the feasibility of training a latent world model that, given sub-goals, can be leveraged to plan actions on a Franka robot arm and perform prehensile manipulation tasks from a monocular RGB camera zero-shot in a new environment.

To summarize, we show that joint-embedding predictive architectures learning from videos can be used to build a world model that enables \emph{understanding} the physical world, \emph{predicting} future states, and effectively \emph{planning} in new situations; this is achieved by leveraging internet-scale video and a small amount of interaction data.
Specifically:
\begin{itemize}

\item \textit{Understanding --- Probe-based Classification:} Scaling self-supervised video pretraining results in video representations applicable to many tasks. \model excels at encoding fine-grained motion information, achieving strong performance on tasks requiring motion understanding, such as Something-Something~v2, with $77.3$ top-1 accuracy using an attentive probe.

\item \textit{Understanding --- Video Question-Answering:} \model encoder can be used to train a multi-modal large language model, to tackle video-question answering tasks. We observe state-of-the-art performance on 8B language model class on multiple benchmarks that require physical world understanding and temporal reasoning, such as MVP ($44.5$ paired accuracy), PerceptionTest ($84.0$ test set accuracy), TempCompass ($76.9$ multi-choice accuracy), TemporalBench ($36.7$ multi-binary short-QA accuracy) and TOMATO ($40.3$ accuracy).  In particular, we show that a video encoder pre-trained \textit{without} language supervision can be aligned with a language model and achieve state-of-the-art performance, contrary to conventional wisdom~\citep{yuan2025tarsier2, wang2024internvideo2}.

\item \textit{Prediction:} Large-scale self-supervised video pretraining enhances prediction capabilities.  \model achieves state-of-the-art performance on the Epic-Kitchens-100 human-action anticipation task using an attentive probe, with $39.7$ recall-at-5, which is a $44$\% relative improvement over the previous best model.

\item \textit{Planning:} We demonstrate that \acModel, obtained by post-training \model with only $62$ hours of unlabeled robot manipulation data from the popular Droid dataset, can be deployed in new environments to solve prehensile manipulation tasks using planning with given subgoals.
Without training on any additional data from robots in our labs, and without any task-specific training or reward, the model successfully handles prehensile manipulation tasks, such as Grasp and Pick-and-Place with novel objects and in new environments.
\end{itemize}

The remainder of this paper is organized as follows. 
\Cref{section:stage1} describes the \model pretraining procedure, including the key ingredients enabling scaling beyond the original V-JEPA recipe of \citet{bardes2024revisiting}.
\Cref{section:stage2} then introduces our approach to training a task-agnostic action-conditioned world model, \acModel, leveraging the pretrained \model model.
\Cref{sec:robot_planning} demonstrates using \acModel for robot control via model-based planning.
Because \acModel models world dynamics in a learned representation space, its capabilities fundamentally depend on the information captured in the \model representation space, and so we further explore the performance of \model for video understanding in \Cref{sec:encoder_comparison} and prediction tasks in \Cref{section:prediction}.
Finally, in \Cref{section:language_understanding} we show that \model can be aligned with a language model for video question answering.
\Cref{section:related_work} discusses related work, and we conclude in \Cref{section:conclusion}.

\begin{figure}[!t]
    \centering
    \begin{subfigure}[T]{0.33\textwidth}
        \centering
        \includegraphics[width=\linewidth]{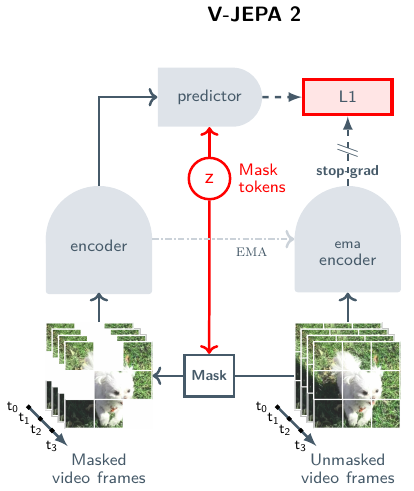}
    \end{subfigure}
    \quad\quad\quad {\color{gray}\vrule} \quad\quad\quad
    \begin{subfigure}[T]{0.33\textwidth}
        \centering
        \includegraphics[width=\linewidth]{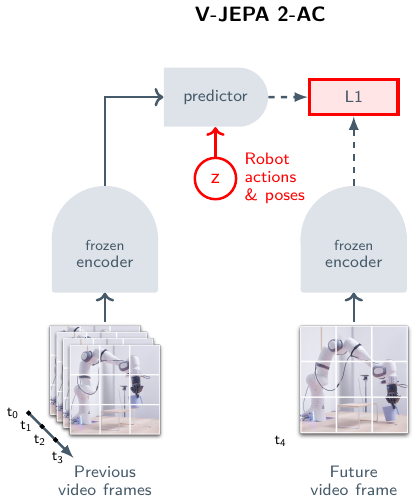}
    \end{subfigure}
    \hfill
    \caption{{\bf Multistage training.}
    {\bf (Left)} We first pretrain the \model video encoder on internet-scale image and video data using a visual mask denoising objective~\citep{bardes2024revisiting, assran2023self}.
    A video clip is patchified into a sequence of tokens and a mask is applied by dropping a subset of the tokens.
    The encoder then processes the masked video sequence and outputs an embedding vector for each input token.
    Next, the outputs of the encoder are concatenated with a set of learnable mask tokens that specify the position of the masked patches, and subsequently processed by the predictor.
    The outputs of the predictor are then regressed to the prediction targets using an L1 loss.
    The prediction targets are computed by an ema-encoder, the weights of which are defined as an exponential moving average of the encoder weights.
    {\bf (Right)} After pretraining, we freeze the video encoder and learn a new action-conditioned predictor, \acModel, on top of the learned representation.
    We leverage an autoregressive feature prediction objective that involves predicting the representations of future video frames conditioned on past video frames, actions, and end-effector states.
    Our action-conditioned predictor uses a block-causal attention pattern such that each patch feature at a given time step can attend to the patch features, actions, and end-effector states from current and previous time steps.
    }
    \label{fig:multistage_training}
\end{figure}

\section{V-JEPA~2: Scaling Self-Supervised Video Pretraining}
\label{section:stage1}

We pretrain \model on a visual dataset that includes over 1 million hours of video. The self-supervised training task is based on mask denoising in representation space and builds upon the V-JEPA framework~\citep{bardes2024revisiting}. In this paper, we extend the V-JEPA framework by exploring larger-scale models, increasing the size of the pretraining data, and introducing a spatial and temporal progressive resolution training strategy that enables us to efficiently pretrain models beyond short 16-frame video clips.

\subsection{Methodology}

\paragraph{\bf Mask-Denoising in Representation Space.}
The V-JEPA objective aims to predict the learned representation of a video $y$ from a view $x$ of that video that has been masked, i.e., from which patches have been randomly dropped (\Cref{fig:multistage_training}, left). The task meta-architecture consists of an encoder, $E_\theta(\cdot)$, which extracts video representations, and a predictor, $P_\phi(\cdot)$, which predicts the representation of masked video parts. The encoder and predictor are trained simultaneously using the objective,
\begin{equation}
\label{eq:loss}
\text{minimize}_{\theta,\phi, \Delta_y}\quad \lVert P_\phi(\Delta_y, E_\theta(x)) - \text{sg}(E_{\overline\theta}(y)) \rVert_1,
\end{equation}
where $\Delta_y$ is a learnable mask token that indicates the locations of the dropped patches. The loss uses a stop-gradient operation, $\text{sg}(\cdot)$, and an exponential moving average, $\overline \theta$, of the weights $\theta$ of the encoder network to prevent representation collapse. The loss is applied only to the predictions of the masked patches.

\paragraph{\bf Architecture.} The encoder, $E_\theta(\cdot)$, and predictor, $P_\phi(\cdot)$, are each parameterized as a vision transformer~\citep{dosovitskiy2020image} (or ViT).
To encode relative position information in the vision transformer, we leverage RoPE (Rotary Position Embedding) instead of the absolute sincos position embedding used in~\citet{bardes2024revisiting}. We use a 3D extension of traditional 1D-RoPE~\citep{su2024roformer} by partitioning the feature dimension into three approximately equal segments (for the temporal, height, and width axes) and applying the 1D rotations separately to the segment for each axis.
We found that using 3D-RoPE instead of absolute sincos position embeddings~\citep{vaswani2017attention} helps stabilize training for the largest models.
To process a video with our transformer encoder, we first patchify it as a sequence of tubelets of size $2\times16\times16$ ($T\times H\times W$) and employ the same multiblock masking strategy as in~\citet{bardes2024revisiting}.

\paragraph{\bf Key Scaling Ingredients.} In this section we introduce and study four additional key ingredients which enable scaling the V-JEPA pre-training principle to obtain our \model model.
\begin{enumerate}
\item \emph{Data scaling:} We increase the dataset size from 2~million to 22~million videos by leveraging and curating additional data sources.

\item \emph{Model scaling:} We scale the encoder architecture from 300~million to over 1~billion parameters, going from a ViT-L to a ViT-g \citep{zhai2022scaling}.

\item \emph{Longer training:} Adopting a warmup-constant-decay learning rate schedule simplifies hyperparameter tuning and enables us to extend training from 90~thousand up to 252~thousand iterations, effectively leveraging the additional data.

\item \emph{Higher resolution:} We leverage the warmup-constant-decay schedule to efficiently scale to higher resolution video and longer video clips by training on shorter, lower-resolution clips during the warmup and constant phases, and then increasing resolution and/or clip-length during the final decay phase.
\end{enumerate}
The remainder of this section describes each of these ingredients in further detail and also quantifies the impact of each ingredient using the evaluation protocol described next.

\begin{wrapfigure}{r}{0.4\textwidth}
    \centering
    \small
    \includegraphics[width=\linewidth]{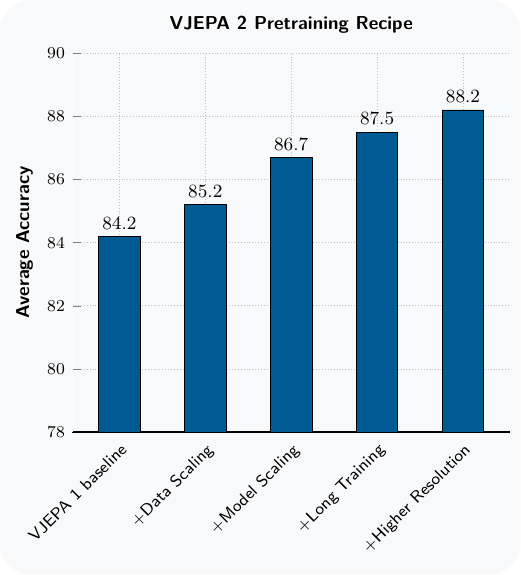}
    \caption{{\bf Scaling Ingredients.} 
    The effects of scaling interventions on average accuracy across 6 image and video classification tasks (SSv2, Diving-48, Jester, Kinetics, COIN, ImageNet) using a ViT-L/16 model as baseline.
    }
    \label{fig:v1_to_v2}
    \vspace{-2em}
\end{wrapfigure}
\paragraph{\bf Evaluation Protocol.} Our goal with model pretraining is to infuse general visual understanding into our encoder. We therefore evaluate our model and data design choices by assessing the quality of the model's learned representation on a set of six motion and appearance classification tasks: Something-Something v2~\citep{goyal2017something}, Diving-48~\citep{li2018resound},  Jester~\citep{materzynska2019jester}, Kinetics~\citep{kay2017kinetics}, COIN~\citep{tang2019coin}, and ImageNet~\citep{deng2009imagenet}. We use a frozen evaluation protocol: we freeze the encoder weights and train a task-specific 4-layers attentive probe on its representation to output a predicted class. In this section, we focus mainly on the average accuracy across the six understanding tasks. Refer to~\Cref{sec:encoder_comparison} for additional details about the tasks, evaluation protocol, and results.

\subsection{Scaling Self-Supervised Video Learning}

We first present a summary of the key findings of our scaling analysis, where we investigate the impact of the four key ingredients on downstream task average performance.
\Cref{fig:v1_to_v2} illustrates the effects of these scaling interventions on average accuracy across 6 classification tasks, using a ViT-L/16 model pretrained on 2 million videos with the V-JEPA objective as our baseline.
Increasing the dataset from 2 million to 22 million videos (VM22M) yields a 1.0-point improvement. Scaling the model from 300 million to 1 billion parameters (ViT-g/16) provides an additional 1.5-point gain. Extending training from 90K to 252K iterations contributes another 0.8-point improvement. Finally, enhancing both spatial resolution ($256\time256 \rightarrow 384\time384$) and temporal duration  ($16 \rightarrow 64$ frames), during both pretraining and evaluation,  boosts performance to 88.2\%, representing a cumulative 4.0-point improvement over the ViT-L/16 baseline. Each individual change provides a positive impact, confirming the potential of scaling in video self-supervised learning (SSL).

\subsection{Pretraining Dataset}
\label{subsec:pretraining_dataset}

Next, we describe the sources of videos and images that make up our pretraining dataset, and our approach to curating the dataset.
\begin{table}[t!]
    \centering
    \small
    \caption{
      \textbf{VideoMix22M (VM22M) Pretraining Dataset.}
      To build our observation pretraining dataset, we combined four different video sources and one image dataset. We use a source-specific sampling probability during training and apply retrieval-based curation on YT1B to reduce noisy content (e.g., cartoon- or clipart-style).}
    \begin{tabular}{ccccccc}
    \toprule
    Source & Samples & Type & Total Hours &  Apply Curation  & Weight\\
    \midrule
    SSv2~\citep{goyal2017something} & 168K & EgoVideo  & 168 & No & 0.056\\
    Kinetics~\citep{carreira2019short} & 733K & ExoVideo  & 614 & No & 0.188\\
    Howto100M~\citep{miech2019howto100m} & 1.1M & ExoVideo  & 134K & No & 0.318\\
    YT-Temporal-1B~\citep{zellers2022merlot} & 19M & ExoVideo  & 1.6M & Yes & 0.188\\
    ImageNet~\citep{deng2009imagenet} & 1M & Images  & n/a & No & 0.250\\
    \end{tabular}
    \label{table:pretraining_dataset}
\end{table}

\paragraph{\bf Scaling Dataset Size.}
We construct a large-scale video dataset by combining publicly available data sources. Using publicly-available sources in this work enables other researchers to reproduce these results. The overall dataset includes ego-centric videos from the Something-Something v2 dataset (SSv2) introduced in~\citet{goyal2017something}, exo-centric action videos from the Kinetics~400, 600, and 700 datasets~\citep{kay2017kinetics, carreira2018short, carreira2019short}, YouTube tutorial videos from HowTo100M~\citep{miech2019howto100m}, and general YouTube videos from YT-Temporal-1B~\citep{zellers2022merlot}, which we refer to as YT1B. We also include images from the ImageNet dataset~\citep{deng2009imagenet} to increase the visual coverage of the pretraining data.
To enable joint image and video pretraining, we duplicate an image temporally and treat it as a 16-frame video where all frames are identical. During training, we sample from each data source with a weighting coefficient that we determined empirically via manual tuning. The resulting dataset, which we refer to as VideoMix22M (or VM22M), consists of 22 million samples. \Cref{table:pretraining_dataset} lists these data sources and their weights.

\begin{figure}[t]
    \centering
    \begin{subfigure}[b]{0.35\textwidth}
        \centering
        \includegraphics[width=\linewidth]{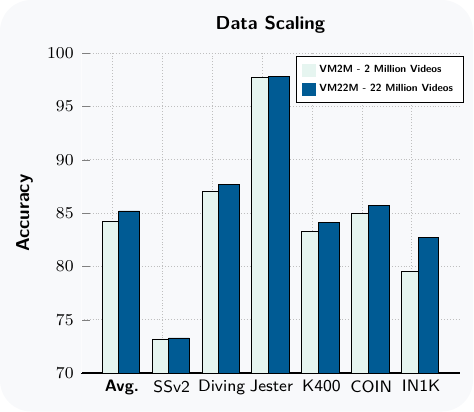}
    \end{subfigure}\hspace{3em}
    \begin{subfigure}[b]{0.35\textwidth}
        \centering
        \includegraphics[width=\linewidth]{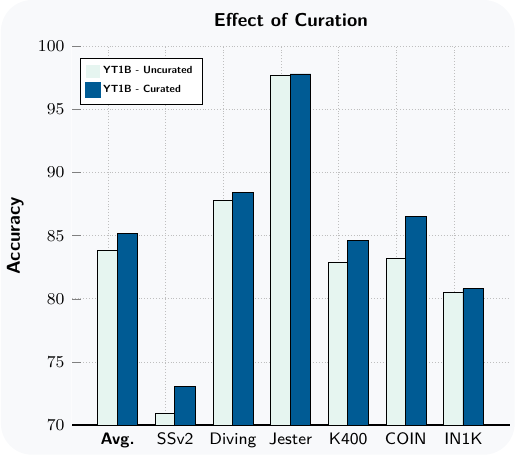}
    \end{subfigure}
    \caption{{\bf Data Scaling \& Curation}. We train and compare models on different data-mixes. Models are ViT-L/16 trained for 90K iterations using a cosine learning schedule following~\citet{bardes2024revisiting}. {\bf (Left)} We compare the performance of a ViT-L/16 model pretrained on the VM2M dataset and our VM22M dataset. Training on the VM22M dataset leads to a $+1$ point improvement in average performance. Performance improvement is more pronounced on appearance-based tasks such as Kinetics-400, COIN, and ImageNet {\bf (Right)} We compare the performance of a ViT-L/16 model pretrained on YT1B and a model pretrained on our Curated-YT1B dataset, which leverages our cluster-based curation. Training on the curated dataset leads to a $+1.4$ point improvement on average performances, showing the effectiveness of data-curation.}

    \label{fig:vjepa2_data_scaling}
\end{figure}

\Cref{fig:vjepa2_data_scaling} (Left) compares the performance of a ViT-L/16 pretrained on VM22M with a similar model trained on the smaller (2 million) VideoMix2M dataset from~\citet{bardes2024revisiting}. Training on VM22M leads to a $+1$ point improvement on average performance on visual understanding tasks, compared to VM2M. Performance improvement is more prominent on appearance-based tasks such as Kinetics-400, COIN, and ImageNet, showing the importance of increasing visual coverage for those tasks.

\paragraph{\bf Data Curation.}
YT1B is a large video dataset, consisting of 1.4 million video-hours, with no curation and minimal filtering compared to smaller video datasets (like Kinetics and Something-Something v2). Because uncurated and unbalanced data can hinder model performance~\citep{assran2022hidden, oquab2023dinov2}, we filter YT1B by adapting an existing retrieval-based curation pipeline to handle videos. Specifically, we  extract scenes from YT1B videos, compute an embedding vector for each scene, and then use a cluster-based retrieval process~\citep{oquab2023dinov2} to select video scenes according to a target distribution, which is composed of the Kinetics, Something-Something v2, COIN and EpicKitchen training datasets. We describe the details of the dataset construction procedure in \Cref{app:pretraining_data_appendix}. Similar to \cite{oquab2023dinov2}, we ensure that none of the videos from the target validation sets are contained in the initial, uncurated data pool.

In \Cref{fig:vjepa2_data_scaling} (Right), we compare the average performance on visual understanding evaluations between a ViT-L model pretrained on uncurated YT-1B data and a comparable model trained on our Curated-YT-1B dataset. Training with the curated dataset yields a $+1.4$ point average performance improvement over the uncurated baseline. Notably, the Curated-YT-1B-trained model achieves competitive performance relative to the full VM22M dataset at the ViT-L scale. However, larger-scale models benefit more from VM22M training (see \Cref{app:pretraining_data_appendix}), suggesting that combining Curated-YT-1B with other data sources enhances scalability.

\subsection{Pretraining Recipe}
\label{subsec:model_training}

\begin{figure}[!t]
    \centering
    \begin{subfigure}[b]{0.32\textwidth}
        \centering
        \includegraphics[width=0.96\linewidth]{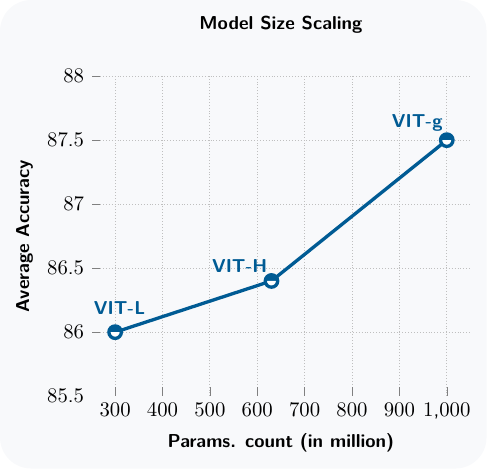}
    \end{subfigure}
    \begin{subfigure}[b]{0.32\textwidth}
        \centering
        \includegraphics[width=\linewidth]{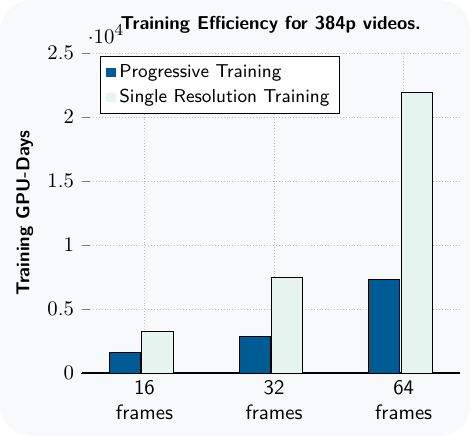}
    \end{subfigure}
    \begin{subfigure}[b]{0.32\textwidth}
        \centering
        \includegraphics[width=\linewidth]{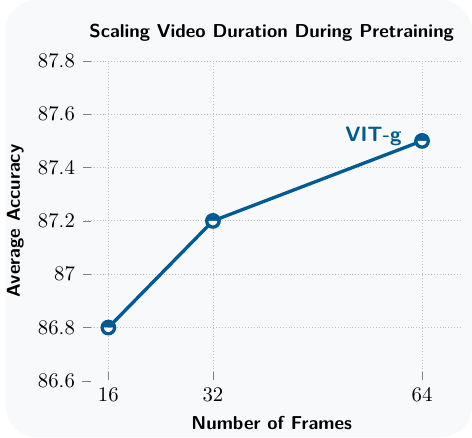}
    \end{subfigure}
    \caption{{\bf Model Scaling.} We explore the impact of scaling model size and input video resolution. All models are trained on the VideoMix22M pretraining dataset.  
{\bf (Left)} Average performance across six understanding tasks as a function of model scale. Models are trained with a constant learning rate until performance plateaus on downstream tasks. We then cool down the model using 64 frames at $256\times256$ resolution and report post-cooldown performance. Scaling the model size from 300M to 1B parameters yields a $+1.7$ point average improvement.  
{\bf (Middle)} Training times (GPU-days) for ViT-g on A100 GPUs when training videos at $384\times384$ resolution with different numbers of frames per clip. We compare progressive resolution training (252K iterations at 16 frames / $256\times256$ resolution, followed by 12K cooldown iterations at $384\times384$ resolution) to the projected time for full-resolution training. Progressive training provides up to 8$\times$ speedup, significantly reducing the pretraining compute requirement.  
{\bf (Right)} Effect of inscreasing video duration at cooldown on downstream performance for ViT-g. Even when only using 16-frame clips during inference/evaluation, increasing video duration during the cooldown phase of training improves average task performance by $+0.7$ points.
}
    \label{fig:vjepa2_scaling}
\end{figure}

\paragraph{\bf Scaling Model Size.}
To explore the scaling behavior of our model, we trained a family of encoder models with parameter counts ranging from 300 million (ViT-L) to 1 billion (ViT-g) parameters. All encoder architecture details are provided in \Cref{table:vit_architecture} in the appendix. Note that each encoder uses the same predictor architecture, similar to a ViT-small.
We report the average performance of these encoders on visual understanding tasks in \Cref{fig:vjepa2_scaling} (Left). Scaling the model size from 300 million (ViT-L) to 1 billion (ViT-g) parameters yields a $+1.5$ points average performance improvement. Both motion and appearance understanding tasks benefit from scaling, with SSv2 improving by $+1.6$ points and Kinetics by $+1.5$ points (cf.\Cref{tb:understanding_main}). These results confirm that self-supervised video pretraining effectively leverages larger model capacities, up to the 1B-parameter ViT-g.

\paragraph{\bf Training Schedule.}
\model model training employs a warmup-constant learning rate schedule followed by a cooldown phase \citep{zhai2022scaling, hagele2024scaling}. Similarly to \citet{hagele2024scaling}, we found that this schedule performs comparably to a half-cosine schedule~\citep{loshchilov2016sgdr}; it also makes exploring long training runs more cost-effective, since multiple cooldown runs can be started from different checkpoints of the constant phase.
We simplified the recipe from \citet{bardes2024revisiting} by maintaining fixed teacher EMA and weight decay coefficients instead of using ramp-up schedule, as these variations showed minimal impact on downstream understanding tasks. \Cref{fig:v1_to_v2} shows that extending the training schedule from 90K to 252K iterations yields a +0.8 average performance improvement with ViT-g models, validating the benefits of extended training durations. This schedule also facilitates progressive training by incrementally increasing video resolution during the cooldown phase.

\paragraph{\bf Efficient Progressive-Resolution Training.}
While most previous video encoders focus on short clips of 16 frames (roughly seconds) \citep{bardes2024revisiting, wang2024internvideo2, wang2023videomae}, we explore training with longer clips of up to 64 frames (16 seconds) at higher spatial resolutions. However, training time increases dramatically with longer durations and higher resolutions --- training our ViT-g model on $64\times384\times384$ inputs would require roughly $60$ GPU-years (see \Cref{fig:vjepa2_scaling}, Middle).
To reduce this, we adopt a progressive resolution strategy~\citep{touvron2019fixing, oquab2023dinov2} that boosts training efficiency while maintaining downstream performance. Our training process begins with a warmup phase where we train on 16-frame, $256\times256$-resolution videos with linear learning rate warmup over 12K iterations, followed by a main training phase with a constant learning rate for 228K iterations. Then, during the cooldown phase, we increase video duration and resolution while linearly decaying the learning rate over 12K iterations. Hence the additional computational overhead associated with training on longer-duration, higher-resolution videos is only incurred during the final cooldown phase. 
This approach enables efficient high-resolution training: as shown in \Cref{fig:vjepa2_scaling} (Middle), we achieve an $8.4\times$ reduction in GPU time for a model that can ingest 64-frame, $384\times384$ resolution inputs, compared to directly training such a model from scratch at full resolution throughout all phases of training. Furthermore, we still observe the benefits of a model that can process longer-duration and higher-resolution inputs as discussed next.

\paragraph{\bf Scaling temporal and spatial video resolution.}
\Cref{fig:vjepa2_scaling} examines how input video resolution affects downstream task performance. When increasing clip duration from 16 to 64 frames during pretraining while maintaining a fixed 16-frame evaluation duration, we observe a $+0.7$ percentage point average performance improvement (\Cref{fig:vjepa2_scaling},  Right). Additionally, we see that increasing the video duration and resolution during evaluation leads to a significant improvement across the tasks (refer to \Cref{tb:understanding_main} and \Cref{eval_frame_duration}).
These results demonstrate that video self-supervised pretraining benefits from increased temporal resolution during both training and evaluation. Although we experimented with scaling to even longer video clips (128 and 256 frames), we did not observe any further improvement beyond 64 frames on this set of understanding tasks. 

\section{\acModel: Learning an Action-Conditioned World Model}
\label{section:stage2}

After pre-training, the \model model can make predictions about missing part in videos. However, these predictions do not directly take into account the causal effect of actions that an agent might take.
In the next stage of training, described in this section, we focus on making the model useful for planning by leveraging a small amount of interaction data.
To that end, we learn a frame-causal action-conditioned predictor on top of the frozen \model video encoder (\Cref{fig:multistage_training}, right). We train our model on data from the Droid dataset~\citep{khazatsky2024droid} consisting of data from experiments with a table-top Franka Panda robot arm collected through teleoperation. We refer to the resulting action-conditioned model as \acModel, and in \Cref{sec:robot_planning} we show that \acModel can be used within a model-predictive control planning loop to plan actions in new environments.

\subsection{Action-Conditioned World Model Training}

Our goal is to take the \model model after pre-training and obtain a latent world model that can be used for control of an embodied agentic system via closed-loop model-predictive control. To achieve this, we train \acModel, an autoregressive model that predicts representations of future video observations conditioned on control actions and proprioceptive observations.

In this section we describe a concrete instantiation of this framework for a tabletop arm with a fixed exocentric camera, and where control actions correspond to end-effector commands.
The model is trained using approximately 62 hours of unlabeled video from the raw Droid dataset, which consists of short videos, typically 3--4 seconds long, of a 7-DoF Franka Emika Panda arm equipped with a two-finger gripper. Here, \emph{unlabeled} video refers to the fact that we do not use additional meta-data indicating any reward, what type of task was being performed in each demonstration, or whether the demonstration was successful or not in completing the task being attempted. Rather, we only use the raw video and end-effector state signals from the dataset (each video in the dataset is accompanied by meta-data indicating the end-effector state in each frame --- three dimensions for position, three for orientation, and one for the gripper state).

\paragraph{\bf Model inputs.}
In each iteration of training we randomly sample a mini-batch of 4 second video clips from the Droid dataset, and, for simplicity, discard any videos shorter than 4 seconds, leaving us with a smaller subset of the dataset comprising under 62 hours of video.
The video clips are sampled with resolution $256\times256$ and a frame-rate of 4 frames-per-second (fps), yielding 16 frame clips denoted by $(x_k)_{k\in[16]}$, where each $x_k$ represents a single video frame.
The robot's end-effector state in each observation is denoted by the sequence $(s_k)_{k\in[16]}$, where $s_k$ is a real-valued 7D vector defined relative to the base of the robot.
The first three dimensions of $s_k$ encode the cartesian position of the end-effector, the next three dimensions encode its orientation in the form of extrinsic Euler angles, and the last dimension encodes the gripper state.
We construct a sequence of actions $(a_k)_{k\in[15]}$ by computing the change in end-effector state between adjacent frames.
Specifically, each action $a_k$ is a real-valued 7-dimensional vector representing the change in end-effector state between frames $k$ and $k+1$.
We apply random-resize-crop augmentations to the sampled video clips with the aspect-ratio sampled in the range (0.75, 1.35).

\paragraph{\bf Loss function.}
We use \model encoder $E(\cdot)$ as an image encoder and encode each frame independently in a given clip to obtain a sequence of feature maps $(z_k)_{k\in[16]}$, where $z_k \coloneqq E(x_k) \in \mathbb{R}^{H \times W \times D}$ with $H \times W$ denoting the spatial resolution of the feature map, and $D$ the embedding dimension.
In practice, our feature maps are encoded using the ViT-g encoder and have the shape $16 \times 16 \times 1408$.
Note that the encoder is kept frozen during this post-training phase.
The sequence of feature maps, end-effector states, and actions are temporally interleaved as $(a_k, s_k, z_k)_{k \in [15]}$ and processed with the transformer predictor network $P_\phi(\cdot)$ to obtain a sequence of next state representation predictions $(\hat{z}_{k+1})_{k \in [15]}$.
The scalar-valued teacher-forcing loss function is finally computed as
\begin{equation}
    \label{eq:loss-wm-tf}
    \mathcal{L}_{\text{teacher-forcing}}(\phi) \coloneqq \frac{1}{T} \sum^{T}_{k=1} \lVert \hat{z}_{k+1} - z_{k+1} \rVert_1 = \frac{1}{T} \sum^{T}_{k=1} \left\lVert P_\phi\left( \left(a_t, s_t, E(x_t)\right)_{t \leq k} \right) - E(x_{k+1}) \right\rVert_1,
\end{equation}
with $T=15$.
We also compute a two-step rollout loss to improve the model's ability to perform autoregressive rollouts at inference time.
For simplicity of exposition and with slight overloading of notation, let $P_\phi(\hat{a}_{1:T}; s_k, z_k) \in \mathbb{R}^{H \times W \times D}$ denote the final predicted state representation obtained by autoregressively running \acModel with an action sequence $(\hat{a}_i)_{i \in [T]}$, starting from ($s_k$, $z_k$).
We can now denote the rollout loss as:
\begin{equation}
    \label{eq:loss-wm-rollout}
    \mathcal{L}_{\text{rollout}}(\phi) \coloneqq \lVert P_\phi(a_{1:T}, s_1, z_1) - z_{T+1} \rVert_1.
\end{equation}
In practice we use $T=2$ for computing the rollout loss, such that we only differentiate the predictor through one recurrent step.

\begin{figure}[!t]
    \centering
    \hfill
    \begin{subfigure}[T]{0.475\textwidth}
        \centering
        \includegraphics[width=\linewidth]{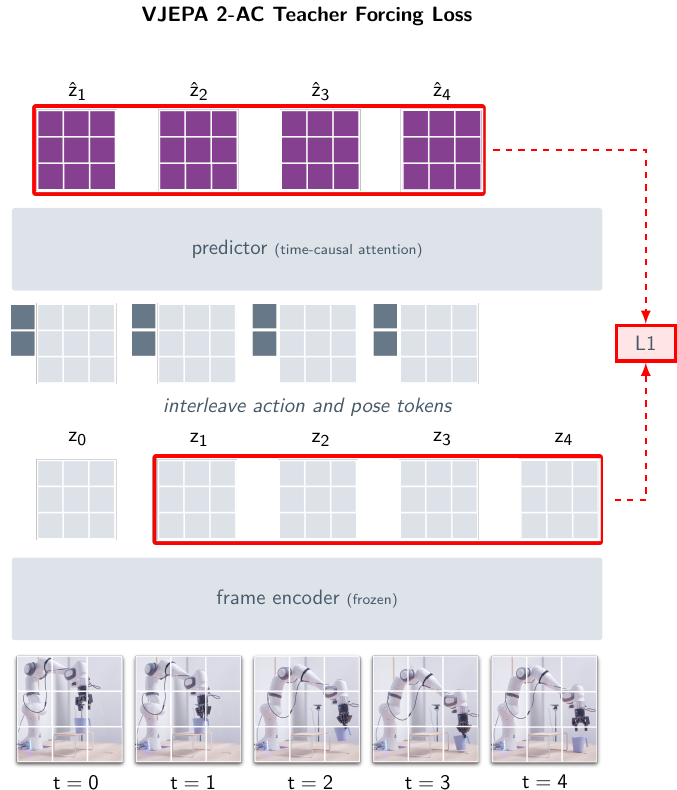}
    \end{subfigure} \hfill {\color{gray}\vrule} \hfill
    \begin{subfigure}[T]{0.475\textwidth}
        \centering
        \includegraphics[width=\linewidth]{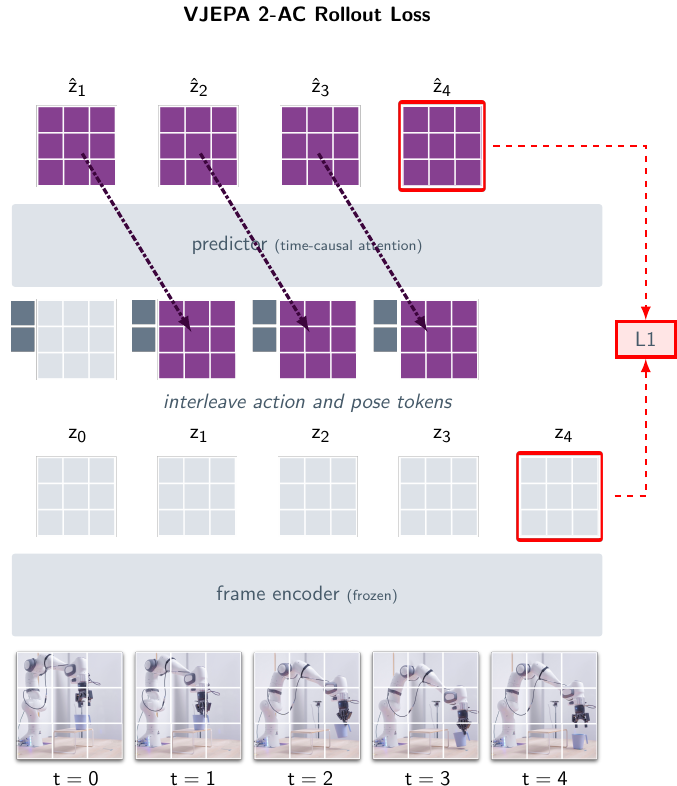}
    \end{subfigure}
    \caption{{\small \bf \acModel training.} \acModel is trained in an autoregressive fashion, utilizing a teacher forcing loss and a rollout loss. ({\bf \small Left}) In the teacher forcing loss, the predictor takes the encoding of the current frame representation as input and learns to predict the representation of the next timestep. ({\bf \small Right}) The rollout loss involves feeding the predictor's output back as input, allowing the model to be trained to predict several timesteps ahead. By optimizing the sum of these two losses, \acModel enhances its ability to accurately forecast the future by reducing error accumulation during rollouts.}
    \label{fig:vjepa2_ac_details}
\end{figure}

The overall training objective is thus given by
\begin{equation}
    L(\phi) \coloneqq \mathcal{L}_{\text{teacher-forcing}}(\phi) + \mathcal{L}_{\text{rollout}}(\phi),
\end{equation}
and is minimized with respect to the predictor weights $\phi$.
For illustrative purposes, the training procedure is depicted in \Cref{fig:vjepa2_ac_details} with $T=4$ for both the teacher forcing and rollout loss.

\paragraph{\bf Architecture.}
The predictor network $P_\phi(\cdot)$ is a $\sim$300M parameter transformer network with 24-layers, 16 heads, 1024 hidden dimension, and GELU activations.
The action, end-effector state, and flattened feature maps input to the predictor are processed with separate learnable affine transformations to map them to the hidden dimension of the predictor.
Similarly, the outputs of the last attention block of the predictor go through a learnable affine transformation to map them back to the embedding dimension of the encoder. 
We use our 3D-RoPE implementation to represent the spatiotemporal position of each video patch in the flattened feature map, while only applying the temporal rotary positional embeddings to the action and pose tokens.
We use a block-causal attention pattern in the predictor so that each patch feature at a given time step can attend to the action, end-effector state, and other patch features from the same timestep, as well as those from previous time steps.

\subsection{Inferring Actions by Planning}

\begin{figure}[!t]
    \centering
    \includegraphics[width=\linewidth]{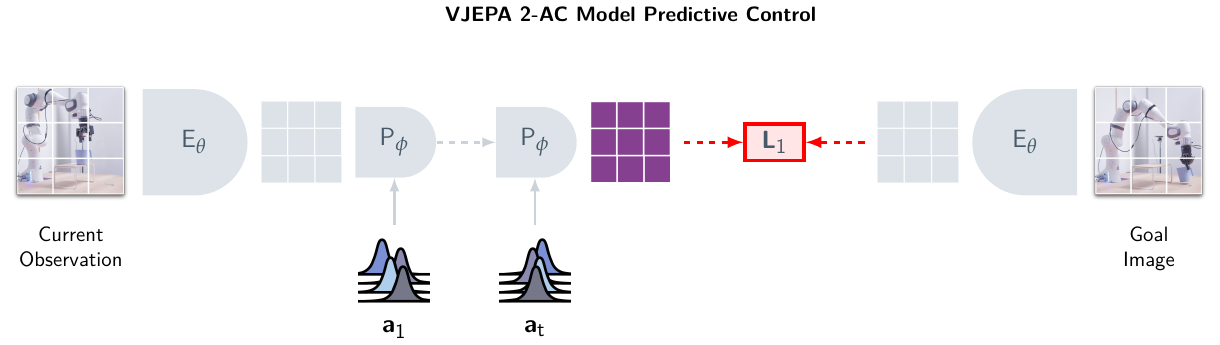}
    \caption{{\bf Planning.}
    We plan an action sequence for a fixed time horizon $T$ by minimizing the L1 distance between the world model's imagined state representation $T$ steps into the future and its goal representation.
    The L1 loss is optimized with respect to the actions $(a_k)_{k \in [T]}$ using the cross-entropy method~\citep{rubinstein1997optimization}.
    Specifically, in each planning step, we sample the action coordinates at each point in the planning horizon from a sequence of Gaussian distributions initialized with zero mean and unit variance.
    The population statistics of the top-k actions trajectories are used to update the mean and variance of the Gaussian distributions.
    This process is repeated for several iterations before finally returning the mean of the sequence of Gaussians as the selected action trajectory.
    }
    \label{fig:planning}
\end{figure}

\paragraph{\bf Energy minimization.}
Given an image of the goal state, we leverage \acModel for downstream tasks by planning.
Specifically, at each time step, we plan an action sequence for a fixed time horizon by minimizing a goal-conditioned energy function.
We then execute the first action, observe the new state, and repeat the process.
Let $s_k$ denote the current end-effector state, and $x_k$ and $x_g$ denote the current observed frame and goal image, respectively, which are separately encoded with the video encoder to obtain the feature maps $z_k$ and $z_g$.
Given a planning horizon, $T$, we optimize a sequence of robot actions, $(a^\star_i)_{i \in [T]}$, by minimizing a goal-conditioned energy function,
\begin{equation}
    \label{eq:energy}
    \mathcal{E}(\hat{a}_{1:T};\ z_k, s_k, z_g) \coloneqq \lVert P(\hat{a}_{1:T}; s_k, z_k) - z_g \rVert_1,
\end{equation}
such that $(a^\star_i)_{i \in [T]} \coloneqq \text{argmin}_{\hat{a}_{1:T}}\ 
 \mathcal{E}(\hat{a}_{1:T};\ z_k, s_k, z_g)$.
As illustrated in \Cref{fig:planning}, the model infers an action sequence $(a^\star_i)_{i \in [T]}$ by selecting a trajectory that minimizes the L1 distance between the world model's imagined state representation $T$ steps into the future and its goal representation.
In practice, we minimize~\eqref{eq:energy} in each planning step using the Cross-Entropy Method~\citep{rubinstein1997optimization}, and only execute the first action on the robot before re-planning, as in receding horizon control.

\section{Planning: Zero-shot Robot Control}
\label{sec:robot_planning}

In this section we demonstrate how \acModel can be used to implement basic robot skills like reaching, grasping, and pick-and-place via model-predictive control.
We focus on tasks with visual goal specification and show that \acModel generalizes zero-shot to new environments.

\subsection{Experimental Setup}
\paragraph{\bf Baselines.}
We compare the performance of \acModel with two baselines, one vision-language-action model trained with behavior cloning, and one video generation-based world model.

The first baseline is based on the Octo video-language-action model that allows for goal-image conditioning~\citep{team2024octo}.
We start from the open-source weights of the \emph{octo-base-1.5} version of the model, which is pretrained on the \emph{Open-X Embodiment} dataset containing over 1M trajectories.\footnote{In comparison, we train \acModel on 23k trajectories from Droid, including successes and failures.} We fine-tune the Octo model with behaviour cloning on the entire Droid dataset using hindsight relabeling~\citep{andrychowicz2017hindsight,ghosh2019learning} with image goals and end-effector states.
In particular, we sample random segments of trajectories from the Droid dataset during training, and uniformly sample goal images up to 20 timesteps forward in the trajectory.
We use the official open-source code for fine-tuning, including all standard Droid optimization hyperparameters, and leverage single side image view inputs at $256\times256$ resolution, a context of two previous frames, and a horizon of 4 future actions.

The second baseline we compare with is based on the Cosmos video generation model~\citep{agarwal2025cosmos}. 
We start with the open-source weights for the action-free Cosmos model (latent diffusion-7B with continuous tokenizer), which was trained on 20M hours of video, and we fine-tune the model on Droid using the officially-released action-conditioned fine-tuning code.\footnote{\url{https://github.com/nvidia-cosmos/cosmos-predict1}}
To improve performance when training on Droid, we (i) lowered the learning rate to match that used in the video-conditioned Cosmos recipe, (ii) removed the dropout in the video conditioning to improve the training dynamics, and (iii) increased the noise level by a factor of $e^2$, as we observed that the model trained with a lower noise factor struggled to leverage the information in the conditioning frame. Although the Cosmos technical report~\citep{agarwal2025cosmos} mentions using world models for planning or model-predictive control as a future application, to the best of our knowledge this is the first reported attempt using Cosmos models for robot control.

\paragraph{\bf Robot deployment.}
All models are deployed zero-shot on Franka Emika Panda arms with RobotiQ grippers, located in two different labs, neither of which appears in the Droid dataset.
Visual input is provided through an uncalibrated low-resolution monocular RGB camera.
The robots use the same exact model weights and inference code, and similar low-level controllers based on operational space control.
We use blocking control for both the \acModel world model and Cosmos world model (i.e., the system waits for the last commanded action to be completed before sending a new action to the controller) and experiment with both blocking and non-blocking control for Octo, and report the best performance across the two options.
When planning with \acModel and Cosmos, we constrain each sampled action to the L1-Ball of radius $0.075$ centered at the origin, which corresponds to a maximum end-effector displacement of approximately 13 cm for each individual action, since large actions are relatively out-of-distribution for the models.
\begin{figure}[t]
    \centering
    \begin{subfigure}[t]{0.325\textwidth}
        \centering
        {\small Start Frame \quad\quad Goal Frame}\\[1ex]
        \includegraphics[width=0.45\linewidth]{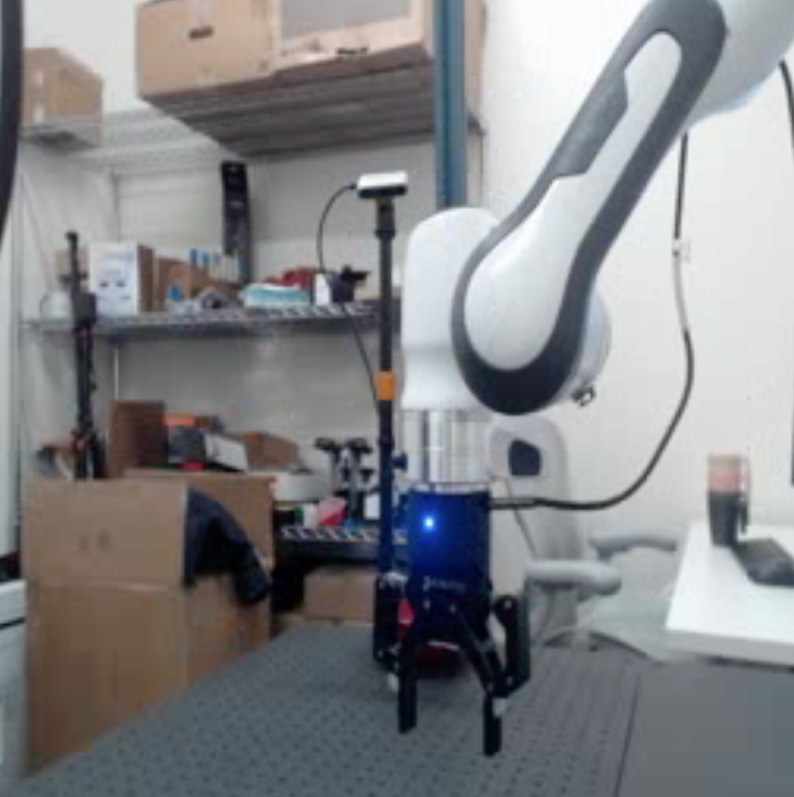}     
        \includegraphics[width=0.45\linewidth]{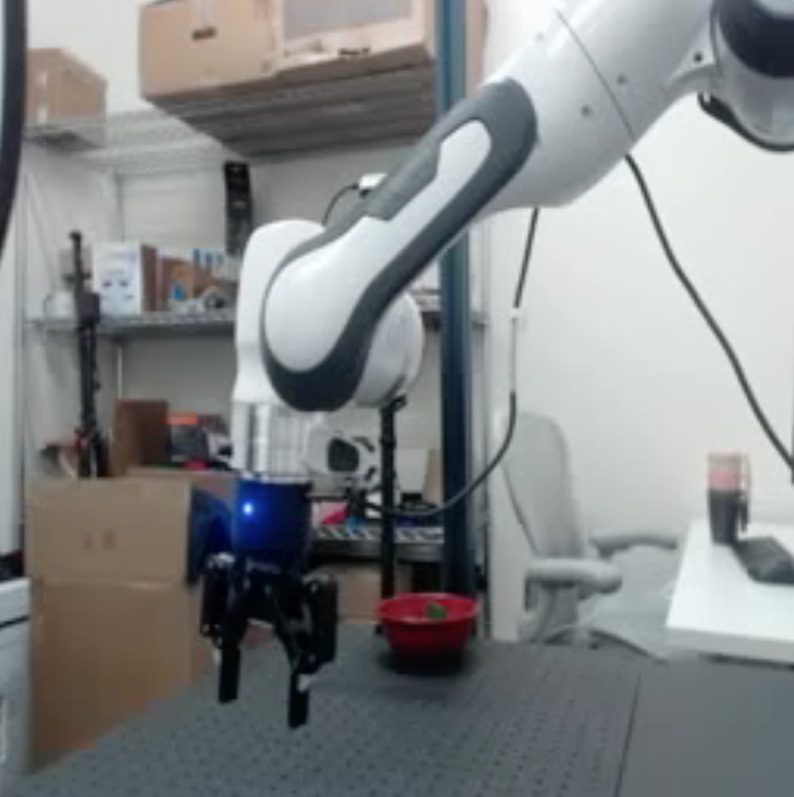}\\[1ex]
        \includegraphics[width=\linewidth]{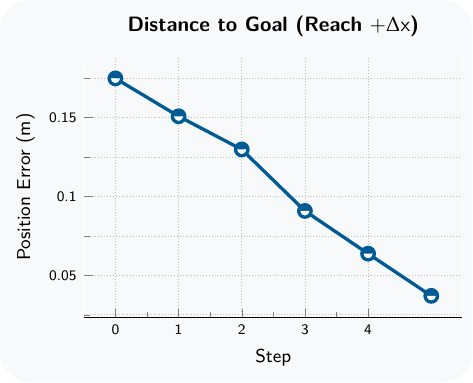}
    \end{subfigure}\hfill
    \begin{subfigure}[t]{0.325\textwidth}
        \centering
        {\small Start Frame \quad\quad Goal Frame}\\[1ex]
        \includegraphics[width=0.45\linewidth]{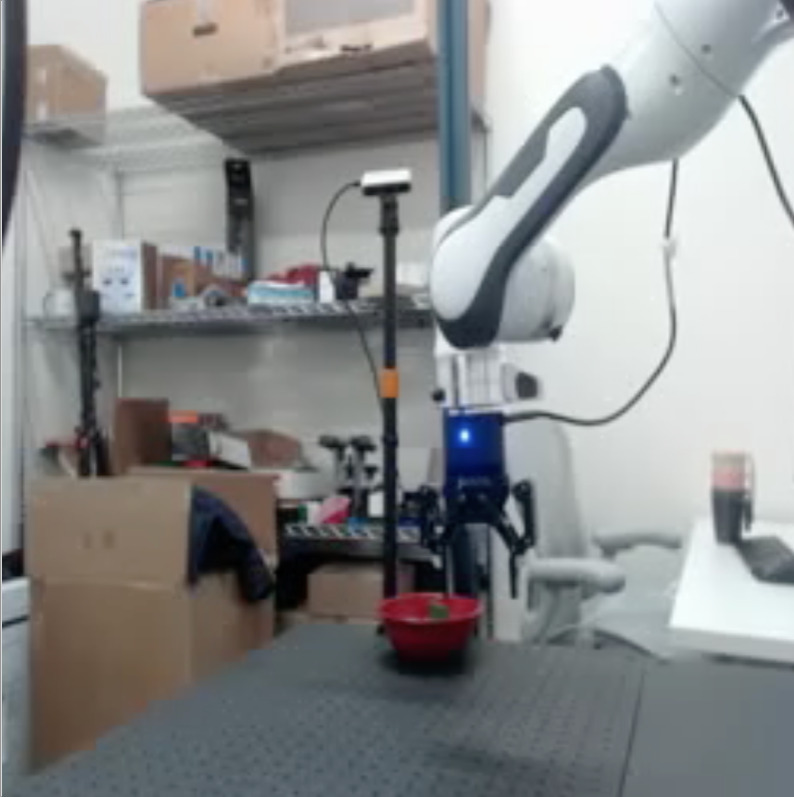}     
        \includegraphics[width=0.45\linewidth]{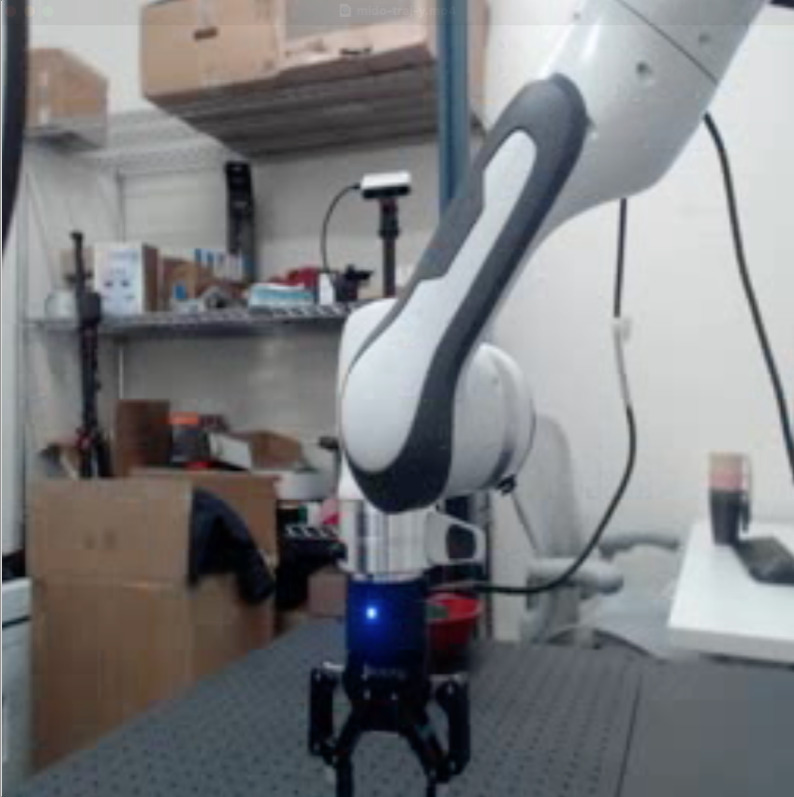}\\[1ex]
        \includegraphics[width=\linewidth]{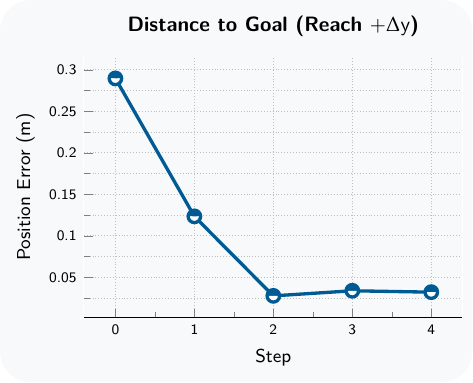}
    \end{subfigure}\hfill
    \begin{subfigure}[t]{0.325\textwidth}
        \centering
        {\small Start Frame \quad\quad Goal Frame}\\[1ex]
        \includegraphics[width=0.45\linewidth]{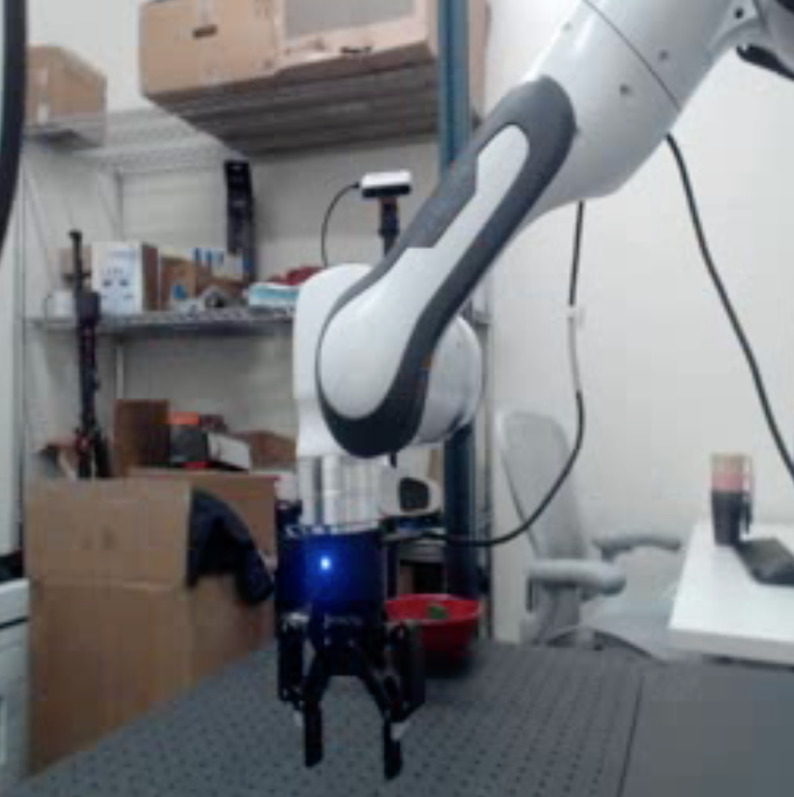}     
        \includegraphics[width=0.45\linewidth]{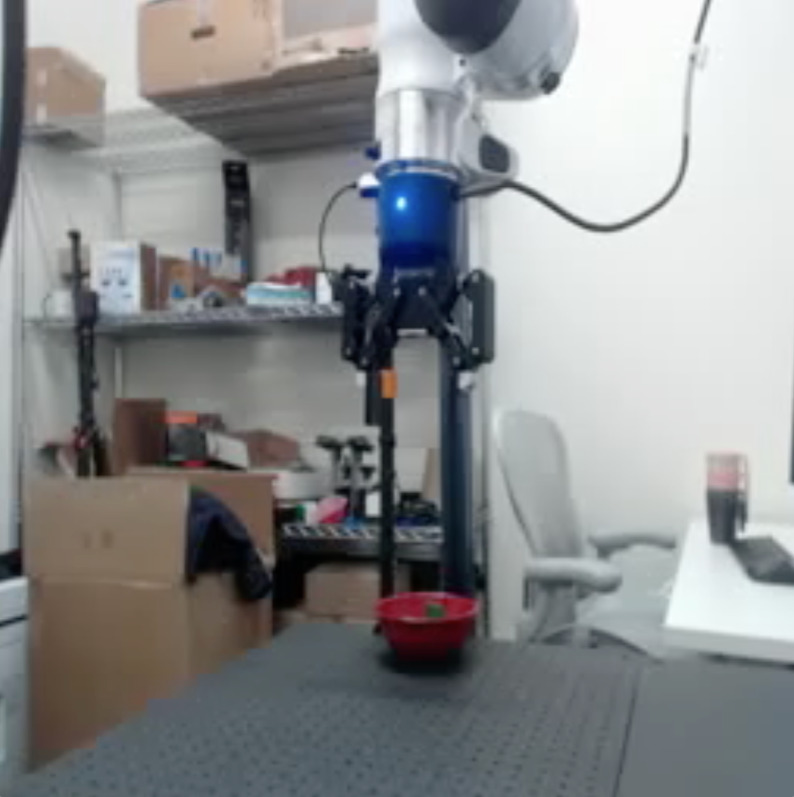}\\[1ex]
        \includegraphics[width=\linewidth]{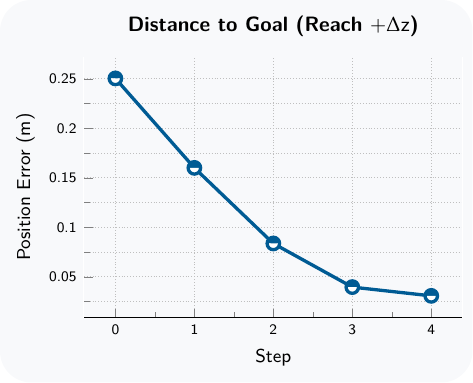}
    \end{subfigure}
    \caption{\small
    {\bf Single-Goal Reaching.} Single-goal reaching involves moving the end-effector to a desired location in space based on a single goal image. This task measures for a basic understanding of actions as well as a 3D spatial understanding of the scene, including depth, from the monocular RGB camera.
    In each step, we use \acModel to plan a sequence of actions by minimizing the L1 distance between the model's imagined future state representations and its representation of the goal frame.
    The first action is then executed before re-planning in the next time step.
    During planning, we only sample individual actions in the L1-Ball of radius $0.075$ centered at the origin.
    Thus, the maximum achievable decrease in cartesian distance to the goal in a single step is $0.13$ ($\sim$13 cm).
    } \label{fig:robot-reach}
\end{figure}

\subsection{Results}

\paragraph{\bf Single-goal reaching.}
First, we evaluate on the task of single-goal reaching, which involves moving the end-effector to a desired location in space based on a single goal image.
This task measures for a basic understanding of actions as well as a 3D spatial understanding of the scene (including depth) from the monocular RGB camera.

\Cref{fig:robot-reach} shows the Euclidean distance between the end-effector and its goal position during robot execution for three different single-goal reaching tasks.
In all cases, the model is able to move the end-effector within less than 4 cm of its goal position, and select actions that lead to a monotonic decrease in the error.
This can be seen as a form of visual servoing~\citep{hill1979real}, wherein visual feedback from a camera is used to control a robot's motion.
However, unlike classical approaches in visual servoing, \acModel achieves this by training on unlabeled, real-world video data.

\begin{wrapfigure}{r}{0.425\textwidth}
    \centering
    \includegraphics[width=\linewidth]{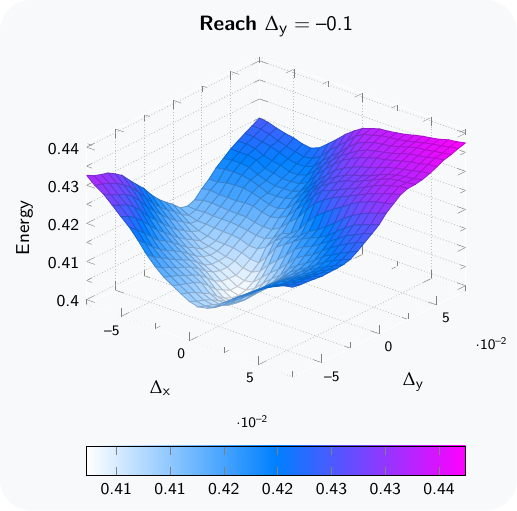}
    \caption{{\bf \acModel Energy Landscape.} Energy landscape for single-goal reaching task with respect to end-effector cartesian-control action (sweeping $\Delta x$ and $\Delta y$ while holding $\Delta z = 0$ fixed); ground truth action relating goal image to start frame is located at $(\Delta x, \Delta y) = (0, -0.1)$.
    We see that the energy function achieves its minimum around $(\Delta x, \Delta y) \approx (0, -0.05)$, indicating that the model has learned to reasonably infer the effect of actions without requiring precision sensing.}
    \label{fig:energy}
\end{wrapfigure}

In \Cref{fig:energy}, we visualize the \acModel energy landscape from equation~\eqref{eq:energy} for the $\Delta y$ reaching task as a function of a single cartesian-control action, sweeping $\Delta x$ and $\Delta y$ while holding $\Delta z = 0$ fixed.
The energy function achieves its minimum near the ground-truth action, providing further evidence that the model has learned to reasonably infer the effect of actions without requiring precision sensing.
It is also interesting to observe that the energy landscape induced by \acModel is relatively smooth and locally convex, which should facilitate planning.

\begin{figure}[t!]
    \centering
    \begin{subfigure}[t]{\textwidth}
        \centering
        \includegraphics[width=\linewidth]{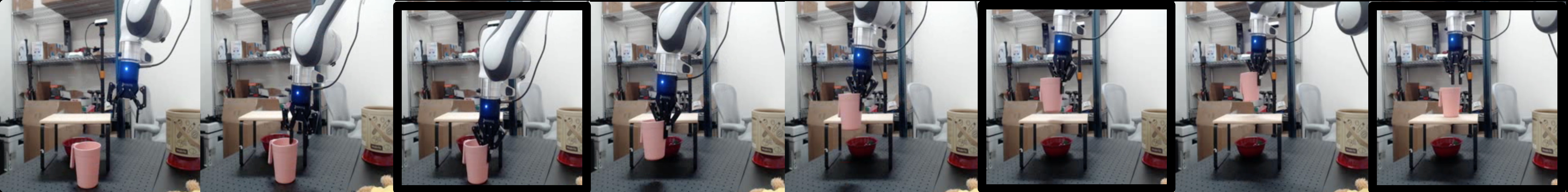}
        \includegraphics[width=\linewidth]{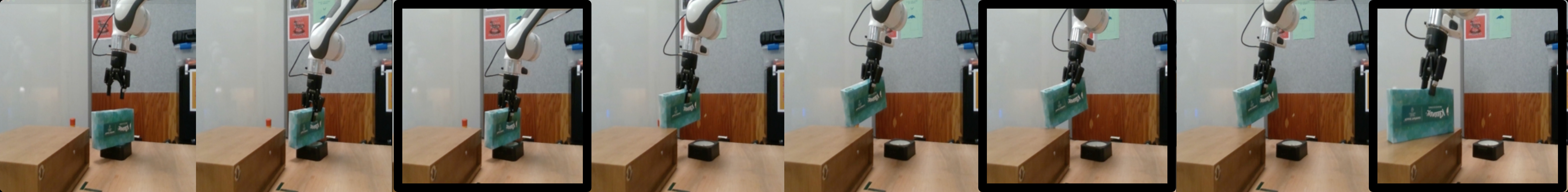}
        \includegraphics[width=\linewidth]{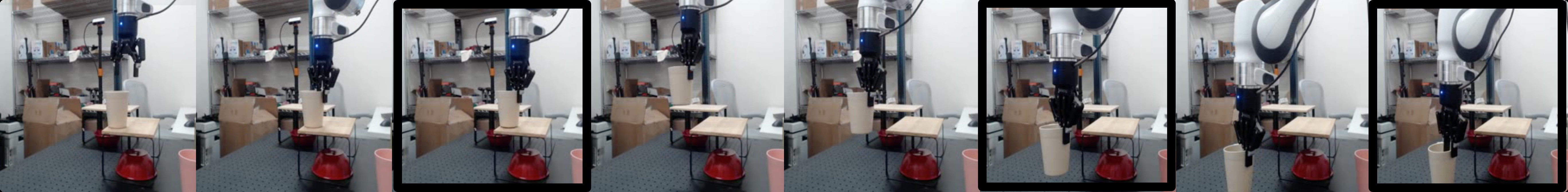}
    \end{subfigure}
    \caption{\small{\bf Pick-\&-Place.} Closed-loop robot execution of \acModel for multi-goal pick-\&-place tasks. 
    Highlighted frames indicate when the model achieves a sub-goal and switches to the next goal.
    The first goal image shows the object being grasped, the second goal image shows the object in the vicinity of the desired location, and the third goal image shows the object placed in the desired position. The model first optimizes actions with respect to the first sub-goal for 4 time-steps before automatically switching to the second sub-goal for the next 10 time-steps, and finally the third goal for the last 4 time-steps.
    Robot actions are inferred through goal-conditioned planning.
    The \acModel model is able to perform zero-shot pick-\&-place tasks on two Franka arms in different labs, with various object configurations and cluttered environments.}
    \label{fig:robot-picknplace}
\end{figure}

\paragraph{\bf Prehensile manipulation.}
Next, we evaluate all models on more challenging prehensile object manipulation tasks, namely \emph{grasp}, \emph{reach with object}, and \emph{pick-and-place}. Success rates are reported in \Cref{tb:robot} and \Cref{tb:cosmos}, and averaged across 10 trials with various permutations to the task across trials (e.g., object location, starting pose, etc.).
For the \emph{grasp} and \emph{reach with object} tasks the model is shown a single goal image.
For the \emph{pick-and-place} tasks we present two sub-goal images to the model in addition to the final goal.
The first goal image shows the object being grasped, the second goal image shows the object in the vicinity of the goal position.
The model first optimizes actions with respect to the first sub-goal for 4 time-steps before automatically switching to the second sub-goal for the next 10 time-steps, and finally the third goal for the last 4 time-steps.
Examples of robot execution for the pick-and-place task are shown in \Cref{fig:robot-picknplace}.
Start and goal frames for all individual tasks in Lab 1 are shown in \Cref{app:robo_exp}.
The \emph{grasp} task requires precise control from visual feedback to correctly grip the object.
The \emph{reach with object} task requires the model to navigate while holding an object, which necessitates a basic understanding of intuitive physics to avoid dropping the object.
Finally, the \emph{pick-and-place} task tests for the ability to compose these atomic skills.

While all models achieve a high success-rate on \emph{reach}, differences in performance are more apparent on tasks involving object interaction.
We observe that the success-rate for all models depends on the type of object being manipulated.
For instance, we find that the cup is mostly easily grasped by placing one finger inside the object and gripping around the rim, however if the control actions produced by the model are not accurate enough, the robot will miss the rim of the cup and fail to grasp the object.
When manipulating the box, there are many more feasible grasping configurations, however, the model requires more precise gripper control to ensure that the fingers are open wide enough to grasp the object.
We see that, for all models, the variation in success-rate with respect to the object type is due to the combination of sub-optimal actions and the unique challenges associated with manipulating each respective object.
Nonetheless, we see that the \acModel model achieves the highest success-rate across all tasks, highlighting the feasibility of latent planning for robot manipulation.

In \Cref{tb:cosmos}, we compare planning performance when using \acModel versus the Cosmos action-conditioned video generation model based on latent diffusion.
In both cases we leverage the cross-entropy method~\citep{rubinstein1997optimization} for optimizing the sequence of actions using a single NVIDIA RTX 4090 GPU, and construct the energy function by encoding the goal frame in the latent space of the model, as in equation~\eqref{eq:energy}.
With 80 samples, 10 refinement steps, and a planning horizon of 1, it takes 4 minutes to compute a single action in each planning step with Cosmos.
While we achieve a high success rate of 80\% on the \emph{reach} tasks when using Cosmos, performance on object interaction tasks is weaker.
Note that under a planning time of 4 minutes per action, a full pick \& place trajectory requires over one hour of robot execution.
By contrast, with 10$\times$ more samples in each refinement step, the \acModel world model requires only 16 seconds per action and leads to higher performance across all considered robot skills.
We can potentially reduce the planning time for both models in future work by leveraging additional computing resources for planning, reducing the number of samples and refinement steps used at each time step, training a feed-froward policy in the world-models' imagination to initialize the planning problem, or potentially leveraging gradient-based planning in the case of \acModel.
    
\begin{table}[t!]
    \centering
    \small
    \caption{\small{\bf Zero-Shot Robot Manipulation.} All models are deployed zero-shot on two Franka arms with RobotiQ grippers located in different labs.
    Given image-goals for each considered task, all models run closed loop to infer a sequence of actions to achieve the goal. Success rates are reported out of 10 trials with various permutations to the task across trials (e.g., object location, starting pose, etc.).}
    \label{tb:robot}
    \begin{tabular}{l c c c c c c c c }
        & & & \multicolumn{2}{c}{\bf Grasp} & \multicolumn{2}{c}{\bf Reach w/ Obj.}
        & \multicolumn{2}{c}{\bf Pick-\&-Place} \\
        \cmidrule(lr{.75em}){4-5} \cmidrule(lr{.75em}){6-7} \cmidrule(lr{.75em}){8-9}
        \bf Method & & \bf Reach & Cup & Box & Cup & Box & Cup & Box \\
        \toprule
        \multirow{3}{*}{{Octo}~{\small\citep{team2024octo}}} & Lab 1 & 100\% & 20\% & 0\% & 20\% & 70\% & 20\% & 10\% \\
         & Lab 2 & 100\% & 10\% & 0\% & 10\% & 70\% & 10\% & 10\% \\
         & \it Avg & \it 100\% & \it 15\% & \it 0\% & \it 15\% & \it 70\% & \it 15\% & \it 10\% \\
        \midrule
        \multirow{3}{*}{\acModel (ours)} & Lab 1 & 100\% & 70\% & 30\% & 90\% & 80\% & 80\% & 80\% \\
        & Lab 2 & 100\% & 60\% & 20\% & 60\% & 70\% & 80\% & 50\% \\
        & \it Avg & \it 100\% & \it 65\% & \it 25\% & \it 75\% & \it 75\% & \it 80\% & \it 65\% \\
        \bottomrule
    \end{tabular}
\end{table}
\begin{table}[t]
    \small
    \caption{
    \small{\bf Planning Performance.} 
    Comparing closed-loop robot manipulation using MPC with \acModel world model and Cosmos world model.
    In both cases we leverage the cross-entropy method~\citep{rubinstein1997optimization} for optimizing the sequence of actions using a single NVIDIA RTX 4090 GPU.
    For each robot skill, we evaluate each model across 10 tasks and average the results.
    With 80 samples, 10 refinement steps, and a planning horizon of 1, it takes 4 minutes to compute a single action in each planning step with Cosmos, which is an action-conditioned video generation model based on latent diffusion.
    Note that under a planning time of 4 minutes per action, a full pick \& place trajectory takes over one hour.
    By contrast, with 10$\times$ more samples in each refinement step, the \acModel world model requires only 16 seconds per action and leads to higher performance across all considered robot skills.  
    }
    \label{tb:cosmos}
    \centering
    \begin{tabular}{lcccc|ccccc}
        Lab 2 & \multicolumn{4}{c}{\bf Planning Details} & & \multicolumn{2}{c}{\bf Grasp} & \multicolumn{2}{c}{\bf Pick-\&-Place} \\
        \cmidrule(lr{.75em}){7-8} \cmidrule(lr{.75em}){9-10}
        \bf Method & \#Samples & Iter. & Horizon & Time & \bf Reach & Cup & Box & Cup & Box \\\midrule
        {Cosmos}~{\small\citep{agarwal2025cosmos}} & 80 & 10 & 1 & 4 min. & 80\% & 0\% & 20\% & 0\% & 0\% \\
        \acModel (ours) & 800 & 10 & 1 & 16 sec. & 100\% & 60\% & 20\% & 80\% & 50\% \\  \bottomrule
    \end{tabular}
\end{table} 
\subsection{Limitations}

\paragraph{\bf Sensitivity to camera positioning.}
Since the \acModel model is trained to predict representations of the next video frame given an end-effector Cartesian control action, without any explicit camera calibration, it must therefore implicitly infer the action coordinate axis from the monocular RGB camera input.
However, in many cases, the robot base is not visible in the camera frame, and thus the problem of inferring the action coordinate axis is not well defined, leading to errors in the world model.
In practice, we manually tried different camera positions before settling on one that worked well across all of our experiments.
We conduct a quantitative analysis of the \acModel world model's sensitivity to camera position in \Cref{subsec:cameraposition}.

\paragraph{\bf Long horizon planning.}
Long horizon planning with world models is limited by a number of factors.
First, autoregressive prediction suffers from error accumulation: the accuracy of the representation-space predictions decreases with longer autoregressive rollouts, thereby making it more difficult to reliably plan over long horizons.
Second, long-horizon planning increases the size of the search space: the number of possible action trajectories increases exponentially given a linear increase in the planning horizon, thereby making it computationally challenging to plan over long horizons.
On the other hand, long-horizon planning is necessary for solving non-greedy prediction tasks, e.g., pick-and-place without image sub-goals.
Future work exploring world models for long-horizon planning will enable the solution of many more complex and interesting tasks.

\paragraph{\bf Image goals.}
Following many previous works in goal-conditioned robot manipulation~\citep{finn2017deep, lynch2020learning, chebotar2021actionable, jang2022bc, liu2022masked, gupta2022maskvit}, our current formulation of the optimization target assumes that we have access to visual goals.
However, when deploying robots in-the-wild, it may be more natural to express goals in other forms, such as with language.
Future work that aligns latent action-conditioned world models with language models will step towards more general task specification via natural language.

\section{Understanding: Probe-based Classification}
\label{sec:encoder_comparison}

The capabilities of a representation-space world model, such as \acModel discussed above, are inherently limited by the state information encoded in the learned representation space. In this section and subsequent sections, we probe the representations learned by \model and compare the \model encoder to other vision encoders on visual classification.

Visual classification tasks can focus either on \emph{appearance understanding} or \emph{motion understanding}. While appearance understanding tasks can generally be solved using information visible in a single frame of an input video clip (even when the classification labels describe actions), motion understanding tasks require several frames to correctly classify a video~\citep{goyal2017something}. 
To ensure a balanced evaluation of both motion and appearance, we have selected three motion understanding tasks, namely Something-Something v2 (SSv2), Diving-48, and Jester, which require the model to understand human gestures and movements. For appearance understanding, we have chosen Kinetics400 (K400), COIN, and ImageNet (IN1K), which involve recognizing actions, scenes, and objects.
Empirically, we show that \model outperforms state-of-the-art visual encoders on motion understanding tasks, while being competitive on appearance understanding tasks.

\paragraph{\bf Attentive Probe.}
We train an 4-layers attentive probe on top of the frozen encoder output using the training data from each task. Our attentive probe is composed of four transformer blocks, the last of which replaces standard self-attention with a cross-attention layer using a learnable query token. Following standard practice, several clips with a fixed number of frames are sampled from a video during inference. The classification logits are then averaged across clips. We keep the resolution similar to the one used for \model pretraining. We ablate the number of layers of our attentive probe in \Cref{app:probe_understanding_add_res}, and also provide full details on the number of clips, clip size, and other hyperparameters used in the downstream tasks.

\begin{table}[t]
  \centering
    \small
    \caption{\small{\bf Action and Object Classification.} We report the classification performance of \model models pretrained on 64 frames at resolution $256\times256$ for all models, except \model ViT-g$_{384}$ which was pretrained at resolution $384\times384$, on action and object classification, and compare their performance with state-of-art image and video encoders. All models follow the same evaluation protocol except for \model ViT-g$_{384}$. We use $256\times256$ resolution with $16\times2\times3$ inputs for SSv2 (16 frames clip, 2 temporal crops, 3 spatial crops), $16\times8\times3$ for K400, $32\times8\times3$ for COIN and $32\times4\times3$ for Diving-48 and Jester. \model ViT-g$_{384}$ uses a higher resolution of $384\times384$ for all six tasks, and additionally uses $64\times2\times3$ inputs for SSv2 and 32x8x3 inputs for COIN. Our \model ViT-g significantly outperforms other vision encoders on motion understanding tasks and is competitive on appearance tasks. It achieves the best average performance of $87.5$ across all image and videos encoders. \model ViT-g$_{384}$ further improves results consistently across tasks, reaching $88.2$ average performance.
      $*$: {\bf PE\textsubscript{core}G} achieves an accuracy $89.8\%$ on ImageNet using an attentive probe and input resoluton of $448$px~\citep{bolya_perception_encoder_2025}. We use an input resolution of $256$px and a different probe architecture in our case.
      }
    \label{tb:understanding_main}
    \vspace{-0.5mm}
    \resizebox{\textwidth}{!}{\begin{tabular}{l c c |c c c | c c c}
      & & & \multicolumn{3}{c|}{\bf\it Motion Understanding} & \multicolumn{3}{c}{\bf\it Appearance Understanding} \\
        \bf Method & \bf Param. & \bf Avg. & \bf SSv2  & \bf Diving-48  & \bf Jester &  \bf K400 & \bf COIN & \bf IN1K \\
        \toprule
        \multicolumn{6}{l}{\bf\it Results Reported in the Literature}\\[1ex]
        {\bf VideoMAEv2}~{\tiny\citep{wang2023videomae}} & 1B & -- &56.1 & -- & -- & 82.8 & -- & 71.4 \\
        {\bf InternVideo2-1B}~{\tiny\citep{wang2024internvideo2}} & 1B & --  & 67.3 & -- & -- & 87.9 & -- & -- \\
        {\bf InternVideo2-6B}~{\tiny\citep{wang2024internvideo2}} & 6B & -- & 67.7 & -- & -- & 88.8 & -- & -- \\
        {\bf VideoPrism}~{\tiny\citep{zhao2024videoprism}} & 1B & -- & 68.5 &  71.3 & -- & 87.6 & -- & -- \\
        \midrule
        \multicolumn{6}{l}{\bf\it Image Encoders Evaluated Using the Same Protocol }\\[1ex]
        {\bf DINOv2}{\tiny~\citep{darcet2024vision}} & 1.1B & 81.1 & 50.7 & 82.5 & 93.4 & 83.6 & 90.7 & 86.1 \\

        {\bf PE\textsubscript{core}G}~{\tiny\citep{bolya_perception_encoder_2025}} & 1.9B & 82.3 & 55.4 & 76.9 & 90.0 & 88.5 & \bf 95.3 & 87.6$^*$\\
        {\bf SigLIP2}{\tiny~\citep{tschannen2025siglip}} & 1.2B & 81.1 & 49.9 & 75.3 & 91.0 & 87.3 & 95.1 & \bf 88.0 \\
        \midrule
        \multicolumn{6}{l}{\bf\it Video Encoders Evaluated Using the Same Protocol }\\[1ex]
        {\bf \oldModel ViT-H}~{\tiny\citep{bardes2024revisiting}}  & 600M & 85.2 & 74.3 & 87.9 & 97.7 & 84.5 & 87.1 & 80.0 \\
        {\bf InternVideo2$_{s2}$-1B}~{\tiny\citep{wang2024internvideo2}} & 1B & 87.0 & 69.7 & 86.4 & 97.0 & \bf 89.4 & 93.8 & 85.8 \\
        \midrule
        \bf \model ViT-L  & 300M & 86.0 & 73.7 & 89.0 & 97.6 & 85.1 & 86.8 & 83.5 \\
        \bf \model ViT-H  & 600M & 86.4 & 74.0 & 89.8 & 97.7 & 85.3 & 87.9 & 83.8 \\
        \bf \model ViT-g  & 1B &  87.5 & 75.3 & 90.1 & 97.7 & 86.6 & 90.7 & 84.6 \\
        \bf \model ViT-g$_{384}$  & 1B & \bf 88.2  &\bf 77.3 & \bf 90.2 & \bf 97.8 & 87.3  & 91.1 & 85.1 \\
\end{tabular}}
\end{table}

\paragraph{\bf Evaluation protocol.}
We compare the performance of \model on motion and appearance tasks with several other visual encoders: DINOv2 with registers~\citep{darcet2024vision} is the current state-of-the-art model for self-supervised learning with images, while SigLIP2~\citep{tschannen2025siglip} and the Perception Encoder PE\textsubscript{core}G~\citep{bolya_perception_encoder_2025} are two state-of-the-art models for image-text contrastive pretraining. We also consider two video encoders: the self-supervised \oldModel~\citep{bardes2024revisiting}, and InternVideo2$_{s2}$-1B~\citep{wang2024internvideo2} which relies primarily on vision-text contrastive pretraining.

We use the same evaluation protocol for every baseline and for \model, learning an attentive probe on top of the frozen encoder, similar to~\citet{bardes2024revisiting}. We adapt image-based models to video following the procedure used in~\citet{oquab2023dinov2}, concatenating the features of each input frame. For InternVideo2$_{s2}$-1B, we use its image positional embedding for the ImageNet task, and for video tasks we interpolate its positional embedding from 4 frames to 8, producing a token count similar to \model. Despite using a common evaluation protocol, the baseline encoders are trained on different data (e.g., DINOv2 on LVD-142M, PE\textsubscript{core}G on MetaCLIP) and are thus not directly comparable. We can therefore only compare different approaches at a system level; i.e., with a consistent evaluation protocol despite differences in training protocol and data. We also include existing results from the literature using a similar frozen protocol, but with potentially different attentive head architecture. In particular, we share reported results of VideoMAEv2~\citep{wang2023videomae}, InternVideo-1B and 6B~\citep{wang2024internvideo2},  and VideoPrism~\citep{zhang2024video} on the classification tasks we consider, when available. We provide complete evaluation and hyperparameters in \Cref{app:probe_understanding_hp}.

\paragraph{\bf Results.}
\Cref{tb:understanding_main} reports the classification performance of \model, the other encoders we evaluated, and other notable results reported in the literature. \model ViT-g (at 256 resolution) significantly outperforms other vision encoders on motion understanding tasks. It achieves a top-1 accuracy of 75.3 on SSv2 compared to 69.7 for InternVideo and 55.4 for PE$_{Core}$G. \model is also competitive on appearance tasks, reaching 84.6 on ImageNet (a $+4.6$ point improvement over \oldModel). Overall, \model obtains the best average performance across all six tasks, compared to other video and image encoders. The higher-resolution, longer-duration \model ViT-g$_{384}$ shows further improvement across all tasks, reaching $88.2$ average performance.

\section{Prediction: Probe-based Action Anticipation}
\label{section:prediction}

\begin{table}[t]
  \centering
    \small
    \caption{\small{\bf Prediction: Human Action Anticipation}. Comparison with the state-of-the-art on the EK100 Action Anticipation benchmark. We report mean-class recall-at-5 for verb, noun and action on the validation set of EK100. \model performance scales linearly with model size and outperforms previous state-of-the-art across all model sizes.}
    \label{tb:prediction_main}
    \begin{tabular}{l c c c c}
      \bf Method & \bf Param. & \multicolumn{3}{c}{\bf\it Action Anticipation}  \\
      & & \small Verb & \small Noun & \small \bf Action \\
        \toprule
        {\bf InAViT}~{\small\citep{roy2024interaction}} & 160M & 51.9 & 52.0 & 25.8 \\
         {\bf Video-LLaMA}~{\small\citep{zhang2023video}}  & 7B & 52.9 & 52.0 & 26.0 \\
        {\bf PlausiVL}~{\small\citep{mittal2024can}}  & 8B & 55.6 & 54.2 & 27.6 \\
        \midrule
        \multicolumn{5}{l}{\bf\it Frozen Backbone}\\[1ex]
        \bf \model ViT-L  & 300M & 57.8 & 53.8 & 32.7 \\
        \bf \model ViT-H & 600M & 59.2 & 54.6 & 36.5 \\
        \bf \model ViT-g & 1B & 61.2 & 55.7 & 38.0 \\
        \bf \model ViT-g$_{384}$ & 1B & \bf 63.6 & \bf 57.1 & \bf 39.7 \\
\end{tabular}
\end{table}

Action anticipation consists in predicting the future action given a contextual video clip leading up to some time before the action. Using the Epic-Kitchens-100 (EK100) benchmark~\citep{Damen2022RESCALING}, we demonstrate that \model action anticipation performance increases consistently with model size. Furthermore, despite only using an attentive probe trained on top of \model representations, we show that \model significantly outperforms prior state-of-the-art approaches that were specifically designed for this task.

\paragraph{\bf Task.}
The EK100 dataset is comprised of 100 hours of cooking activities recorded from an egocentric perspective across 45 kitchen environments. Each video in EK100 is annotated with action segments, which include a start timestamp, an end timestamp, and an action label. There are 3,568 unique action labels, each consisting of a verb and a noun category, with a total of 97 verb categories and 300 noun categories. The EK100 action anticipation task involves predicting noun, verb, and action (i.e., predicting verb and noun jointly) from a video clip, referred to as context, that occurs before the start timestamp of an action segment. The interval between the end of the context and the beginning of the action segment is the anticipation time, which is set to 1 second by default. Given that different future actions may be possible from a given context, mean-class recall-at-5 is used as the metric to measure performance~\citep{Damen2022RESCALING}.

\paragraph{\bf Anticipation Probe.}
An attentive probe is trained on top of the frozen \model encoder and predictor to anticipate future actions.
Specifically, we sample a video clip that ends 1 second before an action starts. This video context is fed to the \model encoder. The predictor takes the encoder representation, along with the mask tokens corresponding to the frame 1 second into the future, and predicts the representation of the future video frame. The outputs of the predictor and encoder are concatenated along the token dimension and fed to an attentive probe with a similar architecture to those used in \Cref{sec:encoder_comparison}, with the difference being that the anticipation probe's final cross-attention layer learns three query tokens (as opposed to one), and each query output is fed to a different linear classifier to predict the action category, the verb category, and the noun category respectively.
A focal loss~\citep{lin2017focal} is applied to each classifier independently and then summed before backpropagating through the shared attention blocks of the probe.
We provide additional details and evaluation hyperparameters in \Cref{app:action_anticipation_hp}.

\paragraph{\bf Baselines.} We compare our model with three baselines that are trained specifically for action anticipation: InAViT~\citep{roy2024interaction} is a a supervised approach that leverages explicit hand-object interaction modeling, and Video-LLaMA~\citep{zhang2023video} and PlausiVL~\citep{mittal2024can} are both approaches that leverage a large language model, with up to 7 billion parameters.

\paragraph{\bf Results.} \Cref{tb:prediction_main} summarizes the results on the EK100 action anticipation benchmark. We compare \model ViT-L, ViT-H and ViT-g encoders, increasing parameter count from 300 millions to 1 billion. All three leverage 32 frames with 8 frames per second at resolution $256\times256$ as video context. We also report results of ViT-g$_{384}$ which uses a resolution of $384\times 384$. \model demonstrates a linear scaling behavior with respect to model size, in terms of action prediction recall-at-5. \model ViT-L with $300$ million parameters achieves $32.7$ recall-at-5. Increasing the size of the model to 1 billion parameters leads to a $+5.3$ point improvement with an action recall-at-5 of $38.0$. Furthermore, \model benefits from using a context with higher resolution, and \model ViT-g$_{384}$ at resolution $384\times384$ improves recall-at-5 by an additional $+1.7$ points over the other models using $256 \times 256$ resolution.

\model outperforms the previous state-of-the-art model PlausiVL by a significant margin, even with its 300 million parameters compared to the 8 billion parameters used in PlausiVL. In particular, \model ViT-g$_{384}$ demonstrates a $+12.1$ points improvement over PlausiVL on action recall-at-5, corresponding to a $44\%$ relative improvement.

\begin{figure}[!t]
    \centering
    \begin{subfigure}[b]{0.98\textwidth}
        \centering
        \includegraphics[trim={0.0cm 3.4cm 1.2cm 2.7cm},clip,width=\linewidth]{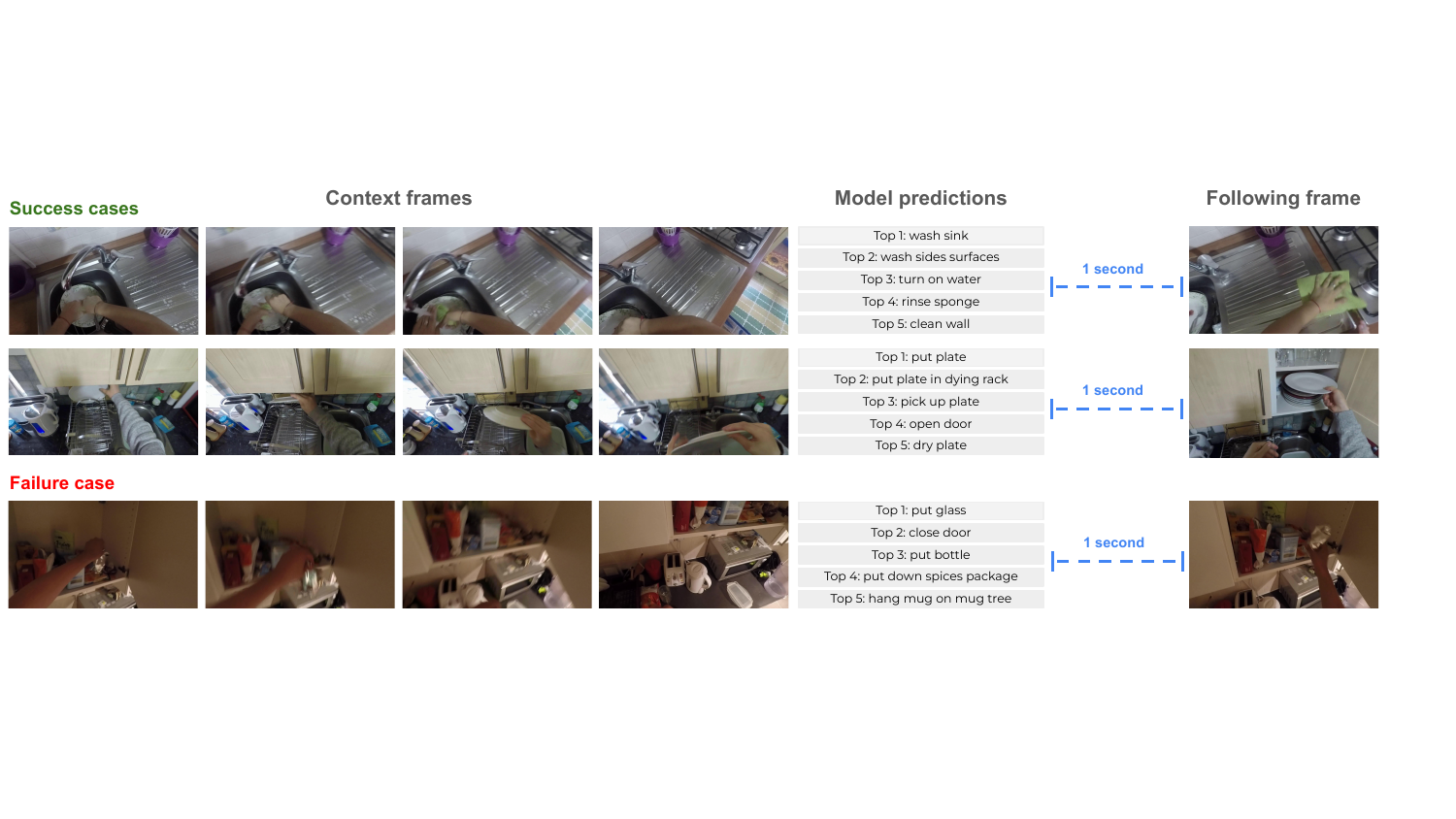}
    \end{subfigure}
    \caption{{\bf Visualization of EK100 prediction.} (Left): four selected frames from the context frames. (Middle): model predictions, ordered by likelihood. (Right): following frame after the 1 second anticipation time. We show two examples where the model is successful and one example where the model fails.}
    \label{fig:ek100_visu}
\end{figure}

In \Cref{fig:ek100_visu} we visualize \model predictions on three samples from the EK100 validation set, two where the model is successful and one where the model fails. For both successful examples, \model not only retrieves the correct action correctly with top 1 confidence, but also proposes coherent top 2 to 5 actions, based on the given context. For example, in the top row, the correct action is "wash sink", but "turn on water" or "clean wall" would both have been valid actions given the presence of a tap and a wall. The model also predicts "rinse sponge", which is the current action being performed, probably assuming that this action could still be going on after 1 second. For the failure case, \model still proposes coherent actions such as "close door" and "put down spices package", but misses the exact nature of the object: "tea package".

\paragraph{\bf Limitations.} \model and the EK100 benchmark have several limitations. First, \model does not fully solve EK100, there are failure cases where the model either gets the verb, the noun, or both wrong. We study the distribution of these failures in \Cref{app:action_anticipation_results}. Second, we focus here on predicting actions with a 1 second anticipation time. The accuracy of \model degrades when predicting at longer time horizons, see \Cref{app:action_anticipation_results}. Third, the EK100 benchmark is limited to kitchen environments, with a closed well-defined vocabulary, and we do not know how well \model generalizes to other environments. This limits the utility and applicability of models trained on EK100. 
Lastly, actions in EK100 are chosen from a fixed set of categories, making it impossible to generalize to action categories not present in the training set.

\section{Understanding : Video Question Answering}
\label{section:language_understanding}

In this section, we explore \model's ability to perform open-language video question answering (VidQA). To enable language capabilities, we train a Multimodal Large Language Model (MLLM) using \model as the visual encoder in the \textit{non-tokenized early fusion} \citep{wadekar2024evolution} setup popularized by the LLaVA family of models \citep{li2024llava}. 
In this family of MLLMs, a visual encoder is \textit{aligned} with a large language model by projecting the output patch embeddings of the vision encoder to the input embedding space of the LLM. The MLLM is then trained either end-to-end, or with a frozen vision encoder.
The majority of the encoders used in MLLMs for VidQA are typically image encoders, which are applied independently per-frame for video inputs~\citep{bai2025qwen2, zhang2024llavanext-video}.
Popular instances of such encoders are CLIP~\citep{radford2021learning}, SigLIP~\citep{tschannen2025siglip}, and Perception Encoder~\citep{bolya_perception_encoder_2025}, which are chosen primarily due to their semantic alignment with language, obtained by pretraining with image-caption pairs.
To the best of our knowledge, our work is the first to use a video encoder that is pretrained \textit{without} any language supervision, to train an MLLM for VidQA.

MLLM performance on downstream tasks is also highly dependent on the alignment data. In these experiments we use a dataset of 88.5 million image- and video-text pairs, similar to what was used to train PerceptionLM~\citep{cho2025perceptionlm}. To demonstrate the effectiveness of the \model encoder, first we compare \model with other state-of-the-art vision encoders in a \textit{controlled} data setup in~\Cref{sec:vqa_comparing_image_encoders}, using a subset of 18 million samples. Then, in the same controlled setup, we show that scaling the vision encoder and input resolution size both consistently improve VidQA performance in~\Cref{sec:vqa_scaling_encoder_and_resolution}. Finally, we scale the alignment data, using the full 88.5 million samples to test the limits of language alignment with \model in~\Cref{sec:vqa_sota}.
Our results demonstrate that in a controlled data setup, \model obtains competitive performance on open-ended VidQA tasks compared to other vision encoders. 
Upon scaling the alignment data, \model achieves state-of-the-art performance on several VidQA benchmarks.

\subsection{Experiment Setup}

\paragraph{\bf Video Question Answering Tasks.}
We evaluate on PerceptionTest~\citep{patraucean2023perception}, which assesses model performance across different skills such as memory, abstraction, physics, and semantics. Additionally, we evaluate on the MVP dataset~\citep{benno2024} for physical world understanding, which utilizes a minimal-video pair evaluation framework to mitigate text and appearance biases. We also evaluate on TempCompass, TemporalBench and TOMATO~\citep{liu2024tempcompass,cai2024temporalbench,shangguan2024tomato} to investigate temporal understanding, and memory capabilities of models. Finally, we report results on general understanding ability using MVBench~\citep{li2024mvbench}, which has a bias towards single-frame appearance features~\citep{benno2024,cores2024tvbench}, and TVBench \citep{cores2024tvbench}, which is proposed in the literature as an alternative for general and temporal understanding, mitigating those biases. 

\paragraph{\bf Visual Instruction Tuning.}
To evaluate the \model representations on visual-question answering tasks, we align \model with an LLM using the visual instruction tuning procedure from the LLaVA framework~\citep{liu2024improved}. 
This process involves converting the visual encoder outputs (or visual tokens) into LLM inputs using a learnable \textit{projector} module, which is typically an MLP. 
We train MLLMs through a progressive three-stage process following~\citet{liu2024llavanext}: Stage 1, where we train the projector solely on image captioning data; Stage 2, where we train the full model on large-scale image question answering, and Stage 3, where we further train the model on large-scale video captioning and question answering. Through this staged training approach, the LLM incrementally improves its understanding of visual tokens.
The vision encoder can either be frozen or finetuned along with the rest of the MLLM. We explore both settings as freezing the vision encoder gives a cleaner signal about the quality of the visual features, while finetuning the vision encoder yields better overall performance.
Further details of the visual instruction training are described in \Cref{app:language_alignment}.

\begin{table}[t]
  \centering
    \caption{\small{\bf Comparison between off-the-shelf image encoders and \model in frozen encoder setting.} All experiments use the same LLM backbone (Qwen2-7B-Instruct), data, and training setup with a \textbf{frozen} vision encoder.  PerceptionTest accuracy is reported on the validation  set post SFT.}
    \label{tb:a2a_llm}
     {\fontsize{8pt}{8pt}\selectfont
    \begin{tabular}{l c | c c c c c c c c }
       \bf Method & \tiny \parbox{2cm}{\centering{\bf Params} \\ Enc / LLM} & \tiny Avg. & \tiny \rotatebox[]{90}{\parbox{2cm}{\centering {\bf PerceptionTest} \\ SFT / Acc}} & \tiny \rotatebox[]{90}{\parbox{2cm}{\centering {\bf MVP} \\ Paired-Acc}}  & \tiny \rotatebox[]{90}{\parbox{2cm}{\centering {\bf TempCompass} \\ multi-choice}}  & \tiny \rotatebox[]{90}{\parbox{2cm}{\centering{\bf TemporalBench} \\ (MBA-short QA) }}  &  \tiny \rotatebox[]{90}{\parbox{2cm}{\centering{\bf TVBench} \\ Acc}} & \tiny \rotatebox[]{90}{\parbox{2cm}{\centering{\bf TOMATO} \\ Acc}} & \tiny \rotatebox[]{90}{\parbox{2cm}{\centering{\bf MVBench} \\ Acc}} \\
        \toprule
        \multicolumn{9}{l}{\bf\it Off-the-shelf image encoders}\\[1ex]
         \bf DINOv2 ViT-g$_{518}$ & 1.1B/7B & 45.7 & 67.1 & 22.4 & 62.3 & 26.8 & 47.6 & 32.0 & 61.8 \\
         \bf SigLIP2 ViT-g$_{384}$ & 1.1B/7B & 48.1 & \bf 72.4 & 26.2 & 66.8 & 25.7 & 48.7 & 33.2 & 64.0 \\
         \bf PE ViT-G/14$_{448}$ & 1.9B/7B & 49.1 & 72.3 & 26.7 & 67.0 & 27.5 & 51.6 & 34.0 & 64.7 \\
         \midrule
         \bf \model ViT-g$_{512}$ & 1B/7B & \bf 52.3 & 72.0 & \bf 31.1 & \bf 69.2 & \bf 33.3 & \bf 55.9 & \bf 37.0 & \bf 67.7 \\
         \bottomrule
    \end{tabular}
    \label{tab:llm_a2a_results}
     }
\end{table}

\begin{table}[t]
  \centering
    \caption{\small{\bf Scaling Vision Encoder Size and Resolution.} We scale the vision encoder from 300 million to 1 billion parameters and input resolution from 256 pixels to 512 pixels. All experiments use the same LLM backbone (Qwen2-7B-Instruct), data, and end-to-end training (unfrozen vision encoder) setup. PerceptionTest accuracy is reported on the validation  set post SFT. Increasing \model encoder scale and resolution improve average performances on VidQA tasks.}
    \label{tb:llm_scaling result}
     {\fontsize{8pt}{8pt}\selectfont
    \begin{tabular}{l c | c c c c c c c c }
      \bf Method & \tiny \parbox{2cm}{\centering{\bf Params} \\ Enc / LLM} & \tiny Avg. & \tiny \rotatebox[]{90}{\parbox{2cm}{\centering {\bf PerceptionTest} \\ SFT / Acc}} & \tiny \rotatebox[]{90}{\parbox{2cm}{\centering {\bf MVP} \\ Paired-Acc}}  & \tiny \rotatebox[]{90}{\parbox{2cm}{\centering {\bf TempCompass} \\ multi-choice}}  & \tiny \rotatebox[]{90}{\parbox{2cm}{\centering{\bf TemporalBench} \\ (MBA-short QA) }}  &  \tiny \rotatebox[]{90}{\parbox{2cm}{\centering{\bf TVBench} \\ Acc}} & \tiny \rotatebox[]{90}{\parbox{2cm}{\centering{\bf TOMATO} \\ Acc}} & \tiny \rotatebox[]{90}{\parbox{2cm}{\centering{\bf MVBench} \\ Acc}} \\
        \toprule
        \multicolumn{8}{l}{\bf\it End-to-end Evaluation}\\[1ex]
        \bf \model ViT-L$_{256}$ & 300M/7B & 51.7 & 74.6 & 32.3 & 70.1 & 30.2 & 50.9 & 36.5 & 67.1 \\
         \bf \model ViT-H$_{256}$ & 600M/7B & 52.0 & 74.7 & 30.6 & 70.9 & 29.8 & 54.6 & 35.1 & 68.0 \\
         \bf \model ViT-g$_{256}$ & 1B/7B & 52.3 & 75.5 & 31.9 & 70.7 & 28.3 & 54.2 & 37.3 & 68.3 \\
         \bf \model ViT-g$_{384}$& 1B/7B & 54.0 & 76.5 & 33.0 &  \bf 71.7 & \bf 33.1 & 56.5 & \bf 39.0 & 68.5 \\
         \bf \model ViT-g$_{512}$& 1B/7B & \bf 54.4 & \bf 77.7 & \bf 33.7 & 71.6 & 32.3 & \bf 57.5 & 38.5 & \bf 69.5 \\ \bottomrule
    \end{tabular}
     }
\end{table}

\subsection{Comparing with Image Encoders}
\label{sec:vqa_comparing_image_encoders}

To isolate the contribution of vision encoders to MLLM performance and compare with \model, we introduce a \textit{controlled} setup: we train individual MLLMs with different state-of-the-art encoders using the \textit{same} LLM backbone and training setup.
In this controlled setup, we use Qwen2-7B-Instruct~\citep{yang2024qwen2technicalreport} and freeze the vision encoder.
We use 18 million image and video-text aligned samples.
We first compare \model, pretrained at resolution 512$\times$512 with DINOv2~\citep{oquab2023dinov2}, SigLIP-2~\citep{tschannen2025siglip}, and Perception Encoder~\citep{bolya_perception_encoder_2025}. 

We observe that \model exhibits competitive performance in the frozen setup, outperforming DINOv2, SigLIP, and Perception Encoder (PE) in all of the tested benchmarks (\autoref{tab:llm_a2a_results}) except PerceptionTest where \model slightly underperforms SigLIP and PE. The improvement is especially noticeable on MVP, TemporalBench, and TVBench --- benchmarks that are primarily focused on temporal understanding. 
Additionally, since we only change the vision encoder, we provide evidence that a video encoder trained without language supervision can outperform encoders trained with language supervision, in contrast to conventional wisdom \citep{tong2024cambrian, li2024llava, liu2024nvila, yuan2025tarsier2}. The results also indicate that using a video encoder instead of an image encoder for VidQA improves spatiotemporal understanding, highlighting the need to develop better video encoders.

\subsection{Scaling Vision Encoder Size and Input Resolution}
\label{sec:vqa_scaling_encoder_and_resolution}
Prior work~\citep{fan2025scaling} suggests that scaling the vision encoder and input resolution significantly improves VQA performance for self-supervised image encoders. Thus, we scale \model from 300M to 1B parameters and the input resolution from 256 to 512 pixels, and show the results in \autoref{tb:llm_scaling result}. When increasing vision encoder capacity from 300M to 1B parameters for a fixed input resolution of 256 pixels, we observe improvements of 0.9 points on PerceptionTest, 3.3 points on TVBench, and 1.2 points on MVBench. Additionally, increasing the input resolution to 512 pixels yields further improvements across all downstream tasks, such as an improvement of 2.2 points on PerceptionTest, 4.0 points on TemporalBench, and 3.3 points on TVBench. These results suggest that further scaling the vision encoder and input resolution is a promising direction for improving VidQA performance.

\begin{table}[t]
    \centering
    \small
    \caption{\small{\bf Comparison with state-of-the-art.}
    We use the full 88.5M-sample alignment dataset and train using the same methodology as PLM 8B~\cite{cho2025perceptionlm}, using a Llama 3.1 backbone. We observe significant improvements in downstream evaluations, obtaining state-of-the-art results in the 8B model class. PerceptionTest accuracy is reported on the test set with SFT for \model; all other results are zero-shot.}
    \label{tab:llm_sota}
    {\fontsize{8pt}{8pt}\selectfont
    \begin{tabular}{l c | c c c c c c c c}
      \bf Method & \tiny \parbox{2cm}{\centering{\bf Params} \\ Enc / LLM} & \tiny Avg. & \tiny \rotatebox[]{90}{\parbox{2cm}{\centering {\bf PerceptionTest} \\ Test Acc}} & \tiny \rotatebox[]{90}{\parbox{2cm}{\centering {\bf MVP} \\ Paired-Acc}}  & \tiny \rotatebox[]{90}{\parbox{2cm}{\centering {\bf TempCompass} \\ multi-choice}}  & \tiny \rotatebox[]{90}{\parbox{2cm}{\centering{\bf TemporalBench} \\ (MBA-short QA) }}  & \tiny \rotatebox[]{90}{\parbox{2cm}{\centering{\bf TOMATO} \\ Acc}} & \tiny \rotatebox[]{90}{\parbox{2cm}{\centering{\bf TVBench} \\ Acc}} & \tiny \rotatebox[]{90}{\parbox{2cm}{\centering{\bf MVBench} \\ Acc}} \\
        \toprule
         \multicolumn{8}{l}{\bf\it $\leq 8B$ Video Language Models Results Reported in the Literature }\\[1ex]
         {\bf InternVL-2.5}~{\tiny\citep{chen2024expanding}}  & 300M/7B & 52.1 & 68.9 & 39.9 & 68.3 & 24.3 & 29.4 & 61.6 & 72.6\\
         {\bf Qwen2VL}~{\tiny\citep{wang2024qwen2}}  & 675M/7B & 47.0 & 66.9 & 29.2 & 67.9 & 20.4 & 31.5 & 46.0 & 67.0\\
         {\bf Qwen2.5VL}~{\tiny\citep{bai2025qwen2}}  & 1B/7B & 49.7 & 70.5 & 36.7 & 71.7 & 24.5 & 24.6 & 50.5 & 69.6 \\
         {\bf PLM 8B}~{\tiny{\citep{cho2025perceptionlm}}} & 1B/8B & 56.7 & 82.7 & 39.7 & 72.7 & 28.3 & 33.2 &  \bf{63.5} & \bf{77.1} \\
        \midrule
        \bf \model ViT-g$_{384}$~{\tiny{LLama 3.1 8B}} & 1B/8B & \bf 59.5 & \bf{84.0}  & \bf{44.5} & \bf{76.9} & \bf 36.7 & \bf 40.3 & 60.6 & 73.5\\ \bottomrule
    \end{tabular}
     }
\end{table}

\subsection{Improving the State-of-the-art by Scaling Data} 
\label{sec:vqa_sota}

After developing a better understanding of the capabilities of \model for training an MLLM in the controlled setup, we study the effect of increasing alignment dataset size to improve the state-of-the-art of VidQA.
Step changes on downstream task performance are often achieved by increasing the scale of the training data, as observed by \citet{cho2025perceptionlm}.
To that end, we increase the scale of MLLM training data from 18 million to the full 88.5 million (4.7$\times$). 
While increasing the model resolution helps in downstream performance, it comes with the challenge of accommodating a large number of visual tokens in the LLM input.
We therefore choose \model ViT-g$_{384}$, leading to 288 visual tokens per frame. 
We follow the same recipe as \citet{cho2025perceptionlm} to train \model ViT-g$_{384}$, using Llama 3.1 as the backbone. To simplify the training process, we use an MLP projector without pooling. 
Details on the scaling training setup are described in \Cref{app:language_alignment}.

Scaling the data uniformly improves the downstream benchmark performance, resulting in state-of-the-art results (\autoref{tab:llm_sota}) on multiple benchmarks --- PerceptionTest, MVP, TempCompass, TemporalBench and TOMATO. 
Compared to the current state-of-the-art PerceptionLM 8B \citep{cho2025perceptionlm}, we observe an increase of 1.3 points on accuracy for PerceptionTest test set, 4.8 points on paired accuracy for MVP, 4.2 points on accuracy for TempCompass, 8.4 points on Multi-binary accuracy for short-QA segment for TemporalBench and 7.1 points on accuracy for TOMATO.
\model does not outperform PerceptionLM on TVBench and MVBench, however it still significantly outperforms other related baselines (InternVL 2.5, Qwen2VL and Qwen2.5VL).
These results underscore the need to scale training data for vision-language alignment and provide evidence that an encoder pretrained without language supervision, such as \model, can achieve state-of-the-art results with sufficient scale.

\section{Related Work}
\label{section:related_work}

\paragraph{\bf World models and planning.}
As early as the work of \citet{sutton1981adaptive} and \citet{chatila1985position}, AI researchers have sought to build agents that use internal models of the world --- modeling both dynamics of the world, as well as mapping the static environment --- to enable efficient planning and control. 
Previous work has investigated world models in simulated tasks~\citep{fragkiadaki2015learning,ha2018world,hafner2019learning,hafner2019dream,hansen2022temporal,hansen2023td,hafner2023mastering,schrittwieser2020mastering,samsami2024mastering}, as well as real-world locomotion and manipulation tasks \citep{lee2020learning, nagabandi2020deep, finn2016unsupervised, ebert2017self, ebert2018visual, yen2020experience}. World model approaches either learn predictive models directly in pixel-space \citep{finn2016unsupervised, ebert2017self, ebert2018visual, yen2020experience}, in a learned representation space \citep{watter2015embed, agrawal2016learning, ha2018world, hafner2019learning, nair2022r3m,wu2023daydreamer,tomar2024video,hu2024learning,lancaster2024modem}, or utilizing more structured representation spaces such as keypoint representations \citep{manuelli2020keypoints, das2020model}.
Previous approaches that have demonstrated real world performance on robotics tasks have trained task-specific world models, and they rely on interaction data from the environment in which the robot is deployed. Evaluation is focused on demonstrating performance of world modeling approaches within the explored task space, instead of generalization to new environments or unseen objects.
In this work we train a task-agnostic world model, and demonstrate generalization to new environments and objects.

Some recent works leverage both internet-scale video and interaction data towards training general purpose (task-agnostic) action-conditioned video generation models for autonomous robots ~\citep{bruce2024genie,agarwal2025cosmos,russell2025gaia}. However, thus far these approaches only demonstrate the ability to generate visually valid-looking plans given actions of the robot, but they have not demonstrated the ability to use those models to actually control the robot. 

Other works have explored the integration of generative modeling into policy learning~\citep{vlp2024,wu2023unleashing,zhao2025cot,zhu2025unified,du2023learning,zheng2025flare,rajasegaran2025empirical}. Differently from this line of work, our goal is to leverage a world model through model-predictive control instead of policy learning to avoid the imitation learning phase that requires expert trajectories. Both approaches are orthogonal and could be combined in future works. Closest to our work, \citet{zhou2024dino,sobal2025learning} show that you can learn a world model stage-wise or end-to-end and use it to solve planning tasks zero-shot. While those previous works focus on small-scale planning evaluation, we show that similar principles can be scaled and used to solve real-world robotic tasks.

\paragraph{\bf Vision-Language-Action models for Robotic Control.}
Recent imitation learning approaches in real-world robotic control have made significant progress towards learning policies that show increasingly good generalization capabilities.
This is achieved by leveraging video-languange models that have been pre-trained on internet scale video and text data, which are then fine-tuned (or adapted) to also predict actions by using behavior cloning from expert demonstrations~\citep{driess2023palm-e,brohan2023rt2,black2024pi0,kim2024openvla,bjorck2025gr00t,intelligence2025pi}. Although these approaches show promising generalization results, it is unclear whether they will be able to learn to predict behaviors that were not demonstrated in the training data since they lack an explicit predictive model of the world and do leverage inference-time computation for planning. They require high-quality large scale teleoperation data, and can only utilize successful trajectories. In contrast, we focus on leveraging any interaction data whether it comes from a successful or failed interaction with the environment. 

\paragraph{\bf Vision Foundation Models.}
Video foundation models in computer vision have shown that large-scale observation datasets comprised of images and/or videos can be leveraged to learn generalist vision encoders that perform well along a wide range of downstream tasks using self-supervised learning approaches from images~\citep{grill2020bootstrap,assran2023self,oquab2023dinov2,fan2025scaling}, videos~\citep{bardes2024revisiting,carreira2024scaling,wang2023videomae,rajasegaran2025empirical}, with weak language supervision~\citep{wang2024internvideo2,bolya_perception_encoder_2025}, or a combination thereof~\citep{tschannen2025siglip,fini2024multimodal}. Previous works, however, tend to focus on understanding evaluation using probe-based evaluation or visual question answering tasks after aligning with a large-language model. While such tasks have served to drive progress, it remains an important goal of a visual system to enable an agent to interact with the physical world~\citep{gibson2014ecological}. Beyond results on visual understanding tasks, we investigate how large-scale self-supervised learning from video can enable solving planning tasks in new environments in a zero-shot manner.

\newpage

\section{Conclusion}
\label{section:conclusion}

This study demonstrates how joint-embedding predictive architectures, learning in a self-supervised manner from web-scale data and a small amount of robot interaction data, can yield a world model capable of understanding, predicting, and planning in the physical world. 
\model achieves state-of-art performances on action classification requiring motion understanding and human action anticipation. \model also outperforms previous vision encoders on video questions-answering tasks when aligned with a large-language model. Additionally, post-training an action-conditioned world model, \acModel, using \model's representation, enables successful zero-shot prehensile manipulation tasks, such as Pick-and-Place, with real-world robots. These findings indicate \model is a step towards developing advanced AI systems that can effectively perceive and act in their environment.

\paragraph{\bf Future work.} There are several important avenues for future work to address limitations of \model. First, in this work we have focused on tasks requiring predictions up to roughly 16 seconds into the future. This enables planning for simpler manipulation tasks, like grasp and reach-with-object, from a single goal image. However, to extend this to longer-horizon tasks such as pick-and-place or even more complex tasks, without requiring sub-goals will require further innovations in modeling. Developing approaches for hierarchical models capable of making predictions across multiple spatial and temporal scales, at different levels of abstraction, is a promising direction.

Second, as mentioned in \Cref{sec:robot_planning}, \acModel currently relies upon tasks specified as image goals. Although this may be natural for some tasks, there are other situations where language-based goal specification may be preferable. Extending the \acModel to accept language-based goals, e.g., by having a model that can embed language-based goals into the \acModel representation space, is another important direction for future work. The results described in \Cref{section:language_understanding}, aligning \model with a language model, may serve as a starting point.

Finally, in this work we scaled \model models up to a modest 1B parameters. The results in \Cref{section:stage1} demonstrated consistent performance improvements while scaling to this level. Previous work has investigated scaling vision encoders to as large as 20B parameters~\citep{zhai2022scaling,carreira2024scaling}. Additional work is needed in this direction to develop scalable pre-training recipes that lead to sustained performance improvements with scale.

\section*{Acknowledgements}

We thank
Rob Fergus, Joelle Pineau, Stephane Kasriel, Naila Murray, Mrinal Kalakrishnan, Jitendra Malik, 
Randall Balestriero, Julen Urain,
Gabriel Synnaeve, Michel Meyer, Pascale Fung, Justine Kao, Florian Bordes, Aaron Foss,
Nikhil Gupta, Cody Ohlsen, Kalyan Saladi, Ananya Saxena, Mack Ward, Parth Malani, Shubho Sengupta, Leo Huang,
Kamila Benzina, Rachel Kim,
Ana Paula Kirschner Mofarrej, Alyssa Newcomb, Nisha Deo, Yael Yungster, Kenny Lehmann, Karla Martucci, 
and the PerceptionLM team, including Christoph Feichtenhofer, Andrea Madotto, Tushar Nagarajan, and Piotr Dollar
for their feedback and support of this project.

\clearpage
\newpage
\bibliographystyle{assets/plainnat}
\bibliography{paper}

\clearpage
\newpage
\beginappendix

\section{\model Pretraining}
\label{app:vjepa2_pretraining}

\subsection{Pretraining Hyperparameters}
As detailed in \Cref{subsec:model_training}, our training pipeline consisted of two phases: 1) a constant learning rate phase and 2) a cooldown phase.
For all models, we trained in the first phase until we observed plateauing or diminishing performance on the IN1K, COIN, and SSv2 tasks. At this point, we initiated the cooldown phase.
\begin{table}[h]
    \small
    \caption{\small{\bf Pretraining Hyperparameters.} Common parameters for pretraining large computer vision models. We report these parameters for both the primary training phase and the cooldown phase.}
    \centering
    \label{tab:pretraining_params}
    \begin{tabular}{l c c}
      \bf Parameter & \bf Primary Phase & \bf Cooldown Phase \\
        \toprule
        Number of frames & 16 & 64 \\
        Frames per Second & 4.0 & 4.0 \\
        Crop Size & 256 & [256, 384, 512] \\
        Random Resize Aspect Ratio & [0.75 1.35] & [0.75, 1.35] \\
        Random Resize Scale & [0.3, 1.0] & [0.3, 1.0] \\
        Steps & Variable & 12000 \\
        Warmup Steps & 12000 & N/A \\
        Batch Size (global) & 3072 & 3072 \\
        Starting Learning Rate & 1e-4 & 5.25e-4 \\
        Final Learning Rate & 5.25e-4 & 1e-6  \\
        Weight Decay & 0.04 & 0.04 \\
        EMA & 0.99925 & 0.99925 \\
        Spatial Mask Scale & [0.15, 0.7] & [0.15, 0.7] \\
        Temporal Mask Scale & [1.0, 1.0] & [1.0, 1.0] \\
        Mask Aspect Ratio & [0.75 1.5] & [0.75, 1.5] \\
        Tubelet Size & 2 & 2 \\
        Patch Size & 16 & 16 \\
        \bottomrule
    \end{tabular}
\end{table}

Training in the first phase began with a learning rate warmup for 12,000 steps followed by a constant learning rate for the rest of the phase.
We checked evaluations every 60,000 steps.
The cooldown phase began with a learning rate at 5.25e-4, which was linearly ramped down to the final learning rate.
Throughout both phases, all other hyperparameters were kept constant.

In the cooldown phase we increased the number of frames per clip while keeping the frames-per-second constant, as we saw a substantial benefit from feeding the model more frames (see \Cref{fig:vjepa2_scaling}).
In addition, we also increased the crop size of the model in this phase, which gave a substantial benefit to tasks like IN1K, which goes from 84.6 at a 256 crop to 85.1 at a 384 crop.
Hyperparameters for both phases are summarized in \Cref{tab:pretraining_params}.

\begin{table}[h]
    \small
    \caption{\small{\bf Abbreviated Pretraining Hyperparameters.} Common parameters for pretraining large computer vision models, targeting our abbreviated recipe.}
    \centering
    \label{tab:abbreviated_pretraining_params}
    \begin{tabular}{l c}
      \bf Parameter & \bf Abbreviated Recipe \\
        \toprule
        Number of frames & 16 \\
        Frames per Second & 4.0 \\
        Crop Size & 256 \\
        Random Resize Aspect Ratio & [0.75 1.35] \\
        Random Resize Scale & [0.3, 1.0] \\
        Steps & 90000 \\
        Warmup Steps & 12000 \\
        Batch Size (global) & 3072 \\
        Starting Learning Rate & 2e-4 \\
        Larning Rate & 6.25e-4 \\
        Final Learning Rate & 1e-6 \\
        Starting Weight Decay & 0.04 \\
        Final Weight Decay & 0.4 \\
        Starting EMA & 0.999 \\
        Final EMA & 1.0 \\
        Spatial Mask Scale & [0.15, 0.7] \\
        Temporal Mask Scale & [1.0, 1.0] \\
        Mask Aspect Ratio & [0.75 1.5] \\
        Tubelet Size & 2 \\
        Patch Size & 16 \\
        \bottomrule
    \end{tabular}
\end{table}
Throughout the Appendix, we refer to a ``abbreviated'' training recipe that corresponds to a 90,000-step training following the procedure of~\citet{bardes2024revisiting}.
There are a few key differences with the abbreviated recipe.
The first is the learning rate: the abbreviated recipe begins with a linear warmup followed by a cosine decay.
The second are the schedules for weight decay and EMA, which are linearly ramped from a starting to a final value.
The last is the total number of steps, which is restricted to 90,000.
We use the abbreviated schedule for several of our ablations on data mixtures, as this allows us to interrogate the effects of data curation on a shorter compute budget.

\subsection{Pretraining data}
\label{app:pretraining_data_appendix}

We began curation of YT1B by applying scene extraction via the PySceneDetect library,\footnote{\url{https://github.com/Breakthrough/PySceneDetect}} which splits videos into clips at scene transitions.
We discard scenes shorter than 4 seconds, retaining 316 million scenes.
The DINOv2 ViT-L model is then applied on the middle frame of each clip to extract scene embeddings.
YT1B embeddings are then clustered into 1.5 million clusters, using the same clustering strategy as~\citet{oquab2023dinov2}.
Embeddings are also extracted in the same manner for all videos in the target distribution, then assigned to the closest YT1B cluster.
We only keep those clusters to which at least one target video was assigned---about 210k clusters out of the original 1.5 million. The retained clusters contain 115 million scenes.

Cluster-based retrieval matches the support, but not the weighting, of the target distribution.
We use a weighted sampling scheme to rebalance the data to better match the target distribution.
W sample from the clusters using a weighted sampling strategy: $w_c = \sum_{d=1}^{D} w_d \times \frac{N_{d,c}}{N_d}$
where $w_c$ is the weighting coefficient for the $c$th cluster, $w_d$ is the weighting coefficient for the $d$th target dataset (from \Cref{table:retrieval_stats}, $N_{d,c}$ is the number of samples from the $d$th dataset in the $c$th cluster, $N_d$ is the total number of samples in the $d$th dataset, and $D$ is the total count of target datasets.
We assigned the retrieval weights approximately based on how many scenes were retrieved by each target dataset, with some extra weighting assigned to EpicKitchen.
This gave a final curated dataset with statistics more closely matching those of handcrafted datasets from the literature.
We found that in isolation, using curated YT1B in place of its uncurated counterpart gave much better results on downstream  understanding tasks (see \Cref{fig:vjepa2_data_scaling}).

\begin{table}[h]
    \centering
    \small
    \caption{
      \textbf{Data Curation Statistics.} We summarized the number of extracted scenes and hours of videos across clusters extracted from YT1B. The final line includes duplicates among retrievals of K710, SSv2, COIN, and EpicKitchen.}
    \begin{tabular}{ccccccc}
    \toprule
    Retrieval Target & Cluster Count & Number of Scenes & Retrieval Weight  \\
    \midrule
    Uncurated YT1B & 1.5M & 316M  \\
    \midrule
    K710 & 170k & 100M &  0.7 \\
    SSv2 & 41k & 19M &  0.125 \\
    COIN & 37k & 21M &  0.125 \\
    EpicKitchen & 4k &  13k & 0.05 \\
    \midrule
    Final Curated (includes duplicates) & 210k & 115M
    \end{tabular}
    \label{table:retrieval_stats}
\end{table}
The overall statistics from how many clusters and scenes were retrieved with this strategy are summarized in \Cref{table:retrieval_stats}.
The overall dataset is weighted towards clusters retrieved with K710. This, combined with its retrieval weight of 0.7, gives the overall curated dataset a heavy Kinetics weighting that we saw reflected in K400 performance for our ablation experiments (see \Cref{app:data_curation_app_additional_results}).
As shown in \Cref{table:pretraining_dataset} in the main body, we combined this Curated YT1B with SSv2, Kinetics, HowTo100M, and ImageNet to create our final VM22M dataset.

\subsection{Scaling Model Size}

Details of the model architecture are shown in \Cref{table:vit_architecture}. All models are parameterized as vision transformers~\cite{dosovitskiy2020image}, using the standard $16 \times 16$ patch size. When scaling model size, we increase the encoder from a ViT-L (300M parameters) to a ViT-g (1B parameters), while the predictor size is kept fixed across all pre-training experiments.

\begin{table}[h]
    \centering
    \small
    \caption{
    \textbf{Model architecture details.}  Family of encoders and predictor architectures used during \model pretraining, with some of the major parameters.}
    \begin{tabular}{lccccccc}
    \toprule
    Model & Params & Width & Depth & Heads & MLP & Embedder\\
    \midrule
    \multicolumn{7}{l}{\bf\it Encoders: $E_\theta(\cdot)$}\\[1ex]
    ViT-L & 300M &  1024 & 24 & 16  & 4096  & $2\times16\times16$ strided conv\\
    ViT-H & 600M &  1280 & 32 & 16 &  5120 &  $2\times16\times16$ strided conv\\
    ViT-g & 1B &  1408 &  40 & 22 & 6144 &  $2\times16\times16$ strided conv\\
    \midrule
    \multicolumn{7}{l}{\bf\it Predictor: $P_\phi(\cdot)$}\\[1ex]
    ViT-s & 22M &  384  & 12  & 12 & 1536 & N.A.\\
    \bottomrule
    \end{tabular}
    \label{table:vit_architecture}
\end{table}

\subsection{Additional Results}
\subsubsection{Effect of Data Curation}
\label{app:data_curation_app_additional_results}
\Cref{table:curated_probe_results} shows the results of data curation on a subset of downstream  classification tasks.
For this table, we trained models at the ViT-L and ViT-g scale using the abbreviated training recipe of the original V-JEPA~\citep{bardes2024revisiting}.
When training smaller scale models (ViT-L), training a model on the curated variant of YT1B leads to across-the-board improvements over the uncurated variant.
However, when moving to a mixed data setting (i.e., adding images and hand-selected videos), performance actually drops for a subset of tasks when using curated data, with performance on SSv2 at 72.8 for VM22M (Mixed+Curated YT1B) vs. 73.3 for Mixed+Uncurated YT1B.
In some cases, the model trained with Curated YT1B alone is better than the one with mixed data, such as on the COIN (86.5 vs. 86.25) and K400 (84.6 vs. 83.7) evaluation tasks.
This result is somewhat surprising, as despite including the K710 training data in the Mixed setting, we find that it does not improve performance over Curated YT1B for the K400 evaluation task.
\begin{table}[h]
    \centering
    \small
    \caption{
      \textbf{Effects of Data Curation on Video Understanding.} Results are reported at both the ViT-L and ViT-g model scales. Models at both scales were pretrained using the abbreviated schedule of~\citet{bardes2024revisiting}.}
    \begin{tabular}{lcccc}
    \toprule
    Training Data & IN1K & COIN & SSv2 & K400  \\
    \midrule
    \multicolumn{3}{l}{\bf\it ViT-L} \\
    Uncurated YT1B & 80.6 & 83.2 & 70.9 & 82.9 \\
    Curated YT1B & 80.8 & \bf 86.5 & 73.1 & \bf 84.6 \\
    Mixed+Uncurated YT1B & \bf 82.9 & 86.25 & \bf 73.3 & 83.0 \\
    VM22M  & \bf 82.9 & 86.0 & 72.8 & 83.7 \\
    \midrule
    \multicolumn{3}{l}{\bf\it ViT-g} \\
    Uncurated YT1B & 81.8 & 86.4 & 73.6 & 85.1 \\
    Curated YT1B & 81.7 & 88.4 & 74.8 & \bf 86.5 \\
    Mixed+Uncurated YT1B & 83.7 & 88.5 & 75.5 & 85.9 \\
    VM22M & \bf 83.9 & \bf 89.2 & \bf 75.6 & 86.2 \\
    \end{tabular}
    \label{table:curated_probe_results}
\end{table}

However, this behavior is not constant across scales.
At the ViT-g scale, VM22M (Mixed+Curated YT1B) outperforms Mixed+Uncurated YT1B on all tasks.

\begin{figure}[t!]
  \centering
  \small
  \includegraphics[width=0.45\linewidth]{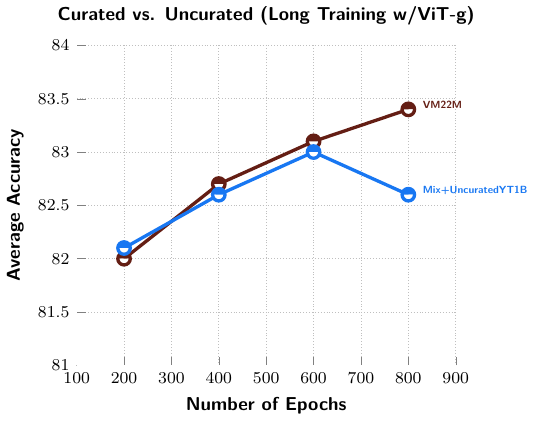}
  \caption{{\bf Effect of data curation for \model pre-training}. We show model performance averaged across the IN1K, COIN, SSv2, and K400 tasks as a function of pre-training ``epochs'' (equivalent to 300 optimization steps). Models trained with and without uncurated data achieve similar performance until epoch 600, at which point the performance of the model trained with uncurated YT1B beings degrading.}
  \label{fig:constlr_data_comparison}
\end{figure}
When following these tasks with the long training schedule, we continue to see differences between VM22M and Mixed+Uncurated YT1B at the ViT-g model scale, as shown in \Cref{fig:constlr_data_comparison}, which compares the performance of the models while averaging across the IN1K, COIN, SSv2, and K400 image understanding tasks.
Initially, the two models improve at roughly the same rate, but their performance diverges after epoch 600 where the model using uncurated data fails to continue improving.

\subsubsection{Effect of Long Training Schedule and cooldown}
\label{eval_frame_duration}
In \Cref{table:effect_of_cooldown} we demonstrate the effects of the two-stage training process.
When comparing to the ViT-g results in \Cref{table:curated_probe_results}, we see that the abbreviated schedule is superior to the constant learning rate schedule prior to the cooldown phase.
The primary benefits are achieved during the cooldown phase, which uses 64 frames for pretraining in combination with a ramped down learning rate.
This leads to a large benefit of over a full point across all evaluations.
\begin{table}[h]
    \centering
    \small
    \caption{
      \textbf{Effects of Long Training and cooldown.} Results are reported at ViT-g model with cooldown at different resolutions.}
    \begin{tabular}{lcccc}
    \toprule
    Training Stage & IN1K & COIN & SSv2 & K400  \\
    \midrule
    Phase 1 (epoch 800, no cooldown) & 83.8 & 89.1 & 75.1 & 85.8 \\
    Phase 2 (annealed, $256\times256$ resolution) & 84.6 & \bf 90.7 & 75.3 & 86.6 \\
    Phase 2 (annealed, $384\times384$ resolution) & \bf 85.1 & 90.2 & \bf 76.5 & \bf 87.3 \\
    \end{tabular}
    \label{table:effect_of_cooldown}
\end{table}

\subsubsection{Effect of Video Length at Evaluation.}

\begin{figure}[t!]
    \centering
    \small
    \includegraphics[width=0.35\linewidth]{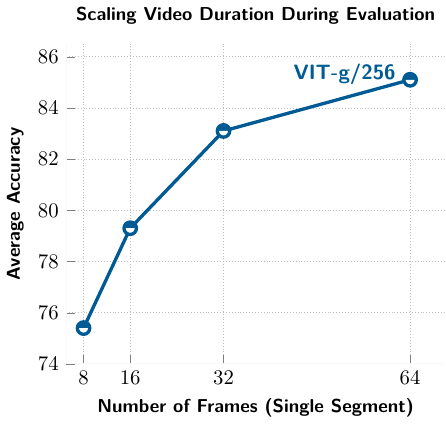}
    \caption{{\bf Effect of video duration during evaluation.} Task performance further improves by running inference on longer video clips. All evaluations use ViT-g models that were annealed with 64 frames at resolution $256\times256$. Due to memory constraints, results are reported using a single clip evaluation protocol. Increasing the number of frames processed at inference time boosts average performance by up to $+9.7$ points.}
    \label{fig:probe_frame_duration}
\end{figure}
\Cref{fig:probe_frame_duration} examines how input video duration affects downstream task performance during evaluation. Using a model pretrained on 64-frame clips, we observe a $+9.7$ percentage point average improvement when increasing the video duration from 16 to 64 frames during evaluation. Note that this ablation uses a single clip evaluation protocol (i.e. we sample only one clip per video) instead of the standard multiclip evaluations due to memory constraints.

\section{\acModel Post-training}
\label{app:vjepa2ac_postraining}

\subsection{Post-Training Hyperparameters}

The \acModel model is trained with the AdamW~\citep{loshchilov2017decoupled} optimizer using a warmup-constant-decay learning-rate schedule, and a constant weight-decay of $0.04$.
We linearly warmup the learning rate from $7.5 \times 10^{-5}$ to $4.25 \times 10^{-4}$ over 4500 iterations, then hold it constant for 85500 iterations, and finally decay it to $0$ over 4500 iterations.
We use a batch size of $256$ comprising 4 second video clips sampled randomly from trajectories in the Droid raw dataset at a frame rate of 4 fps.
We train on the left extrinsic camera views from Droid --- one could also train on videos from right camera views, however we found that training on both left and right camera views, without additionally conditioning on the camera position, degraded performance.
For simplicity, we discard any videos shorter than 4 seconds, leaving us with less than 62 hours of video for training.
We apply random-resize-crop augmentations to the sampled video clips with the aspect-ratio sampled in the range (0.75, 1.35).

\subsection{Robot Task Definitions}
\label{app:robo_exp}
\begin{figure}[h!]
    \centering
    \small
    \begin{subfigure}[b]{\linewidth}
        \includegraphics[width=\linewidth]{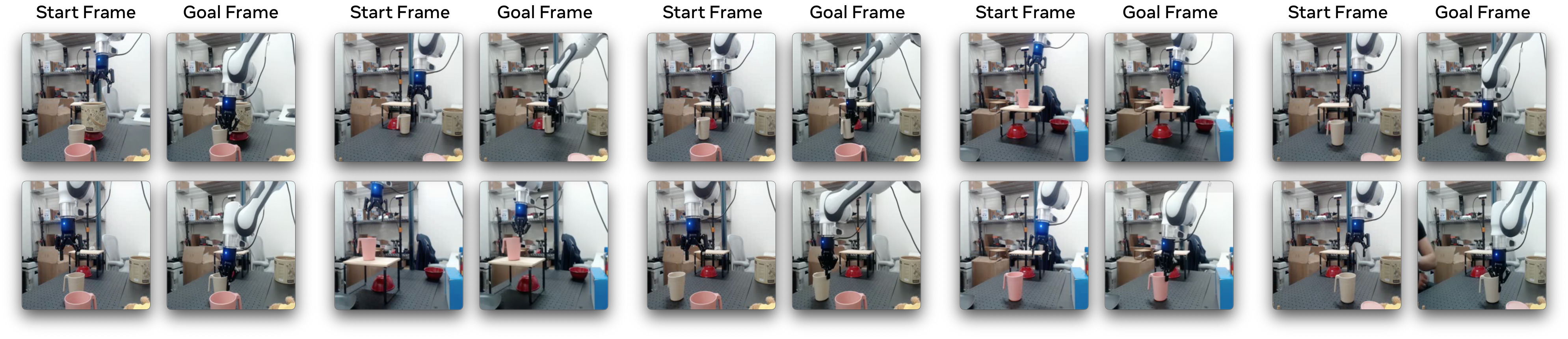}
        \caption{{\bf Grasp Cup}}\vspace{1em}
    \end{subfigure}
    \begin{subfigure}[b]{\linewidth}
        \includegraphics[width=\linewidth]{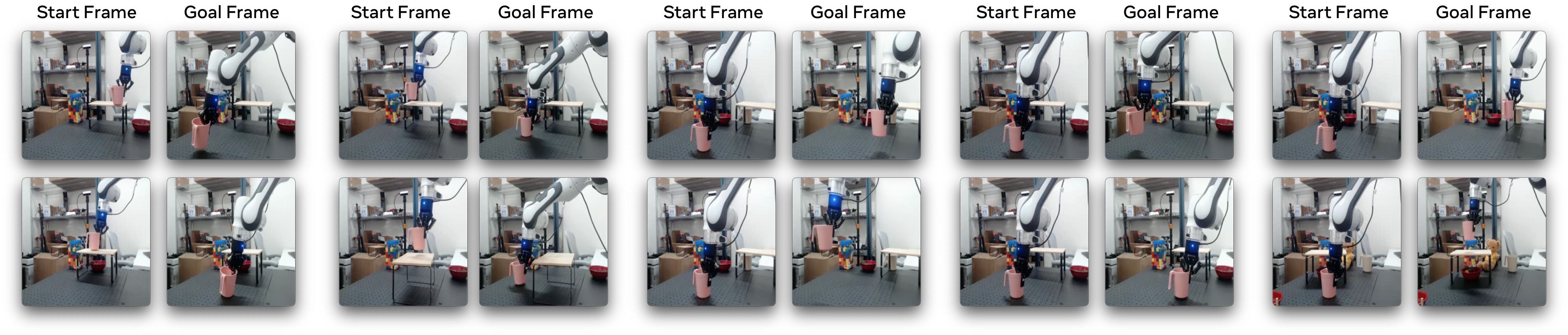}
        \caption{{\bf Reach with Cup}}\vspace{1em}
    \end{subfigure}
    \begin{subfigure}[b]{\linewidth}
        \includegraphics[width=\linewidth]{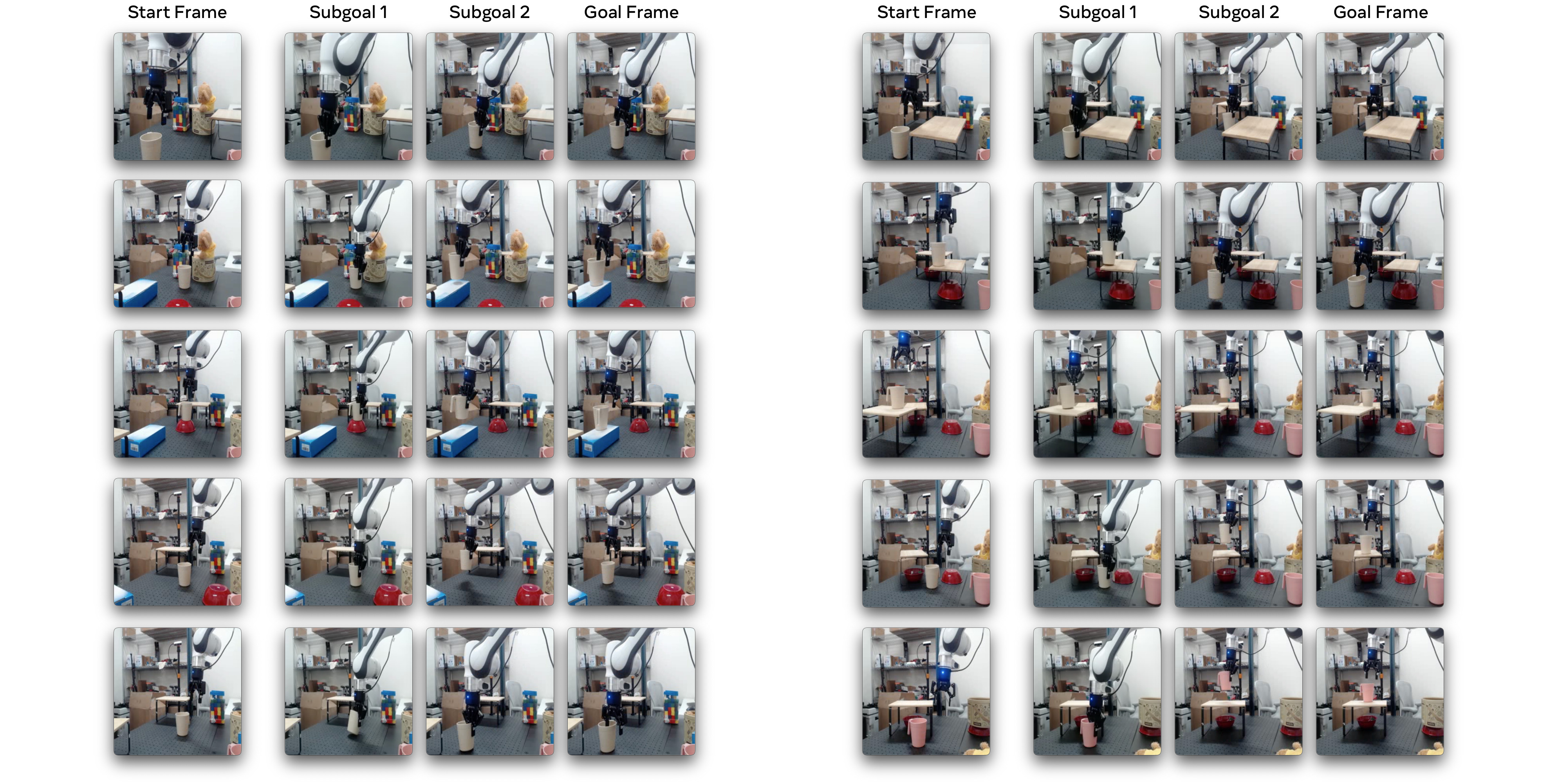}
        \caption{{\bf Pick and Place Cup}}
    \end{subfigure}
    \caption{
    {\bf Prehensile Manipulation Task Definition.}
    Start and goal frames for prehensile manipulation tasks with a cup in Lab 1.
    For the \emph{grasp} and \emph{reach with object} tasks the model is shown a single goal image.
    For the \emph{pick-and-place} tasks we present two sub-goal images to the model in addition to the final goal.
    The first goal image shows the object being grasped, the second goal image shows the object in the vicinity of the goal position.
    The model first optimizes actions with respect to the first sub-goal for 4 time-steps before automatically switching to the second sub-goal for the next 10 time-steps, and finally the third goal for the last 4 time-steps.
    }
    \label{fig:tasks}
\end{figure}

\Cref{fig:tasks} shows examples of start and goal frames for prehensile manipulation task with a cup in Lab 1.
For the \emph{grasp} and \emph{reach with object} tasks the model is shown a single goal image.
For the \emph{pick-and-place} tasks we present two sub-goal images to the model in addition to the final goal.
The first goal image shows the object being grasped, the second goal image shows the object in the vicinity of the goal position.
The model first optimizes actions with respect to the first sub-goal for 4 time-steps before automatically switching to the second sub-goal for the next 10 time-steps, and finally the third goal for the last 4 time-steps.
When planning with \acModel, we use 800 samples, 10 refinement steps based on the top 10 samples from the previous iteration, and a planning horizon of $1$.
Since all considered tasks are relatively greedy, we found a short planning horizon to be sufficient for our setup.
While longer planning horizons also worked reasonably well, they require more planning time.
\afterpage{\clearpage}

\subsection{Visualizing World Model Predictions}
\label{app:extra_exps}

\afterpage{\clearpage}
\begin{figure}[h!]
    \centering
    \small
    \begin{subfigure}[b]{\linewidth}
        \centering
        \begin{minipage}{0.2\textwidth}
            \centering\small
            Robot observations \\[2cm]
            \parbox{2cm}{
                \centering\small
                Reconstructions \\ 
                from \model \\[2cm]
            }
            \parbox{2cm}{
                \centering\small
                Reconstructions \\ 
                from \acModel
            }
        \end{minipage}%
        \begin{minipage}{0.6\textwidth}
            \includegraphics[width=\linewidth]{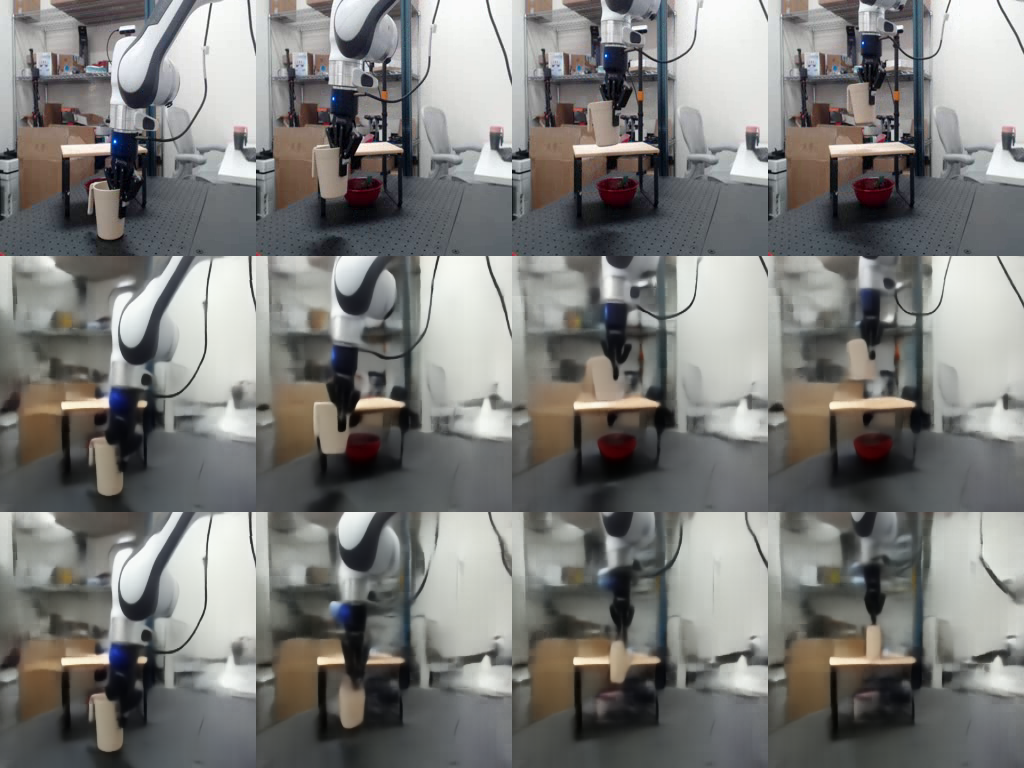}\\
        \end{minipage}
        \caption{{\bf Comparing accuracy of predictions to ground truth trajectory.}
        {\bf (Top Row)} Video frames of a ground-truth trajectory from a robot in our lab.
        {\bf (Middle row)}
        Each frame is encoded by \model encoder, and then decoded using the feedforward frame decoder.
        Reconstructions of the \model representations show that the encoder captures the salient parts of the scene necessary for vision-based control; blurry background generation can be partially attributed to the low-capacity of our feedforward frame decoder.
        {\bf (Bottom Row)}
        Autoregressive rollout produced by \acModel world model using the ground-truth action sequence given first frame as context, and then decoded using feedforward frame decoder.
        Reconstructions of the \acModel rollout show that the action-conditioned world model successfully animates the robot while keeping the background and non-interactated objects (e.g., the shelf) unaffected.
        However, we do observe error accumulation as the world model predicts the location of the cup to be slightly lower than that of the real trajectory in the final frame.}
        \label{fig:wm_decoder}
    \end{subfigure}\vspace{2em}
    \begin{subfigure}[b]{\linewidth}
        \centering
        \begin{minipage}{0.2\textwidth}
            \centering
            \parbox{2cm}{%
                    \centering\small
                    \acModel imagination: move with closed gripper
            }%
            \vspace{1.6cm}
            \parbox{2cm}{%
                    \centering\small
                    \acModel imagination: move with open gripper
            }
        \end{minipage}%
        \begin{minipage}{0.6\textwidth}
            \includegraphics[width=\linewidth]{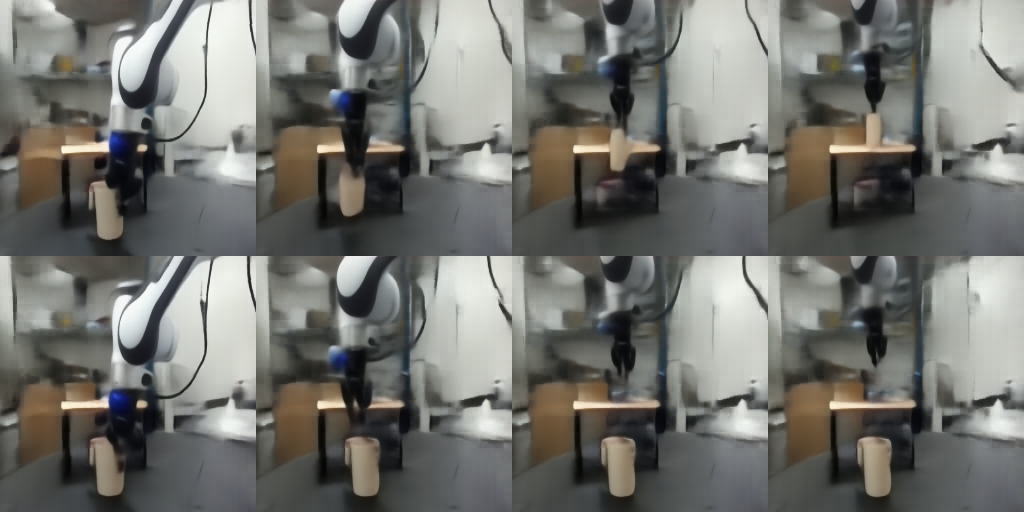}\\
        \end{minipage}
        \caption{
        {\bf Ablating predictions with open versus closed gripper.}
        We explore how the \acModel predictions change when driving the model with identical action sequences, but in one cause using a closed gripper {\bf (top row)} and in the other with an open gripper {\bf (bottom row)}.
        The world model predicts the location of the cup to be unchanged across time steps when using an open gripper action sequence, suggesting a reasonable understanding of intuitive physics (e.g., object constancy, shape constancy, and gravity).
        }
        \label{fig:wm_decoder_contrast}
    \end{subfigure}
    \caption{{\bf Decoding representations.}
    To visualize the model’s predictions, we train a frame decoder on the Droid dataset that maps the \model representations to human-interpretable pixels. The decoder is a feedforward network (fully deterministic regression model that does not use any sampling internally) trained with a mean-squared error pixel reconstruction loss.
    By applying the frame decoder, trained on the \model encoder, to the representations produced by the \acModel predictor, we can visualize world model rollouts for various action sequences.
    }
\end{figure}

To visualize the model's predictions, we train a frame decoder on the Droid dataset that maps the \model representations to human-interpretable pixels.
Specifically, we process 4 frame clips with the frozen \model video encoder,  decode each frame separately using our decoder network, and then update the weights of the decoder using a mean-squared error (L2) pixel reconstruction loss.
The decoder is a feedforward network (fully deterministic regression model that does not use any sampling internally) with output dimension $256 \times 256 \times 3$, parameterized as a ViT-L.
We train the decoder for 150000 optimization steps using AdamW with a fixed weight decay of $0.1$, gradient clipping of $1.0$, and a batch size of $1024$ frames.
We linearly warmup the learning rate for 2000 steps to a peak value of $5 \times 10^{-4}$ and then decay it following a cosine schedule.
For inference, we take the decoder trained on the \model encoder and apply it off-the-shelf to the representations produced by the \acModel predictor.
The decision to only use use a simple feedforward architecture and decode representations at the frame level (as opposed to video level), is to better leverage the decoder as an interpretability tool to analyze the  \acModel rollouts for a set of robot action sequences.

In \Cref{fig:wm_decoder}, we show the video frames of a ground-truth trajectory from a robot in our lab (top row), the decoded \model encoder representations of each frame (middle row), and the decoded \acModel world model rollout using the ground-truth action sequence and a single starting frame as context (bottom row).
Reconstructions of the \model representations (middle row) show that the encoder captures the salient parts of the scene necessary for vision-based control; blurry background generation can be partially attributed to the low-capacity of our feedforward frame decoder.
Reconstructions of the \acModel rollout show that the action-conditioned world model successfully animates the robot while keeping the background and non-interactated objects (e.g., the shelf) unaffected.
We also see that, with a closed gripper, the model correctly predicts the movement of the cup with the arm, suggesting a reasonable understanding of intuitive physics (e.g., object constancy, shape constancy, and gravity), but we do observe error accumulation as the world model predicts the location of the cup to be slightly lower than that of the real trajectory in the final frame.
In \Cref{fig:wm_decoder_contrast} we explore how the \acModel predictions change when driving the model with identical action sequences, but in one cause using a closed gripper (top row) and in the other with an open gripper (bottom row).
The world model predicts the location of the cup to be unchanged across time steps when using an open gripper action sequence.

\subsection{Assessing Sensitivity to Camera Position}
\label{subsec:cameraposition}

\begin{figure}[t!]
    \centering
    \includegraphics[width=\linewidth]{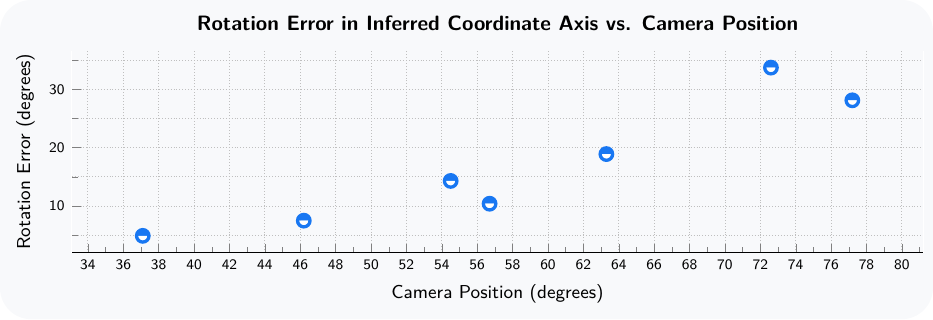}
    \caption{
    {\bf Sensitivity to camera position.}
    Rotation error (in the x-y plane) of the action coordinate axis inferred by \acModel as a function of camera position, with 0 degrees corresponding to a camera located at the robot base, and 90 degrees corresponding to a camera located left of the robot base.
    While ideally, the model's inferred coordinate axis would be invariant to camera position, here we observe that the model's inferred coordinate axis is sensitive to the camera position.
    }
    \label{fig:wm_calibration}
\end{figure}
In practice, we manually tried different camera positions before settling on one that worked best for our experiments; then the camera is kept in the same location for all experiments, across all tasks.
In this section, we conduct a quantitative analysis of the \acModel world model's sensitivity to camera position.
While ideally, the model's inferred coordinate axis would be invariant to camera position, here we observe that the model's inferred coordinate axis is sensitive to the camera position; this is problematic as large errors in the inferred coordinate axis can degrade success rate on downstream tasks.

We sweep several camera positions around the robot base, which we describe as a clockwise angular position around the center of the table, with 0 degrees being located at the robot base, and 90 degrees being left of the robot base.
Since we train on the left exocentric camera views from the Droid dataset, we sweep camera positions between roughly 35 degrees and 85 degrees.
Next, for each camera position, we collect a 201 step trajectory of random robot movements within the horizontal x-y plane.
For each pair of adjacent frames in this 201 step trajectory, we compute the optimal action inferred by \acModel, i.e., the action that minimizes the energy function in eq.~\eqref{eq:energy} given a 1-step rollout.
This allows us to construct a dataset for each camera position consisting of \emph{real action} versus \emph{inferred action} pairs.
We only focus on the $\Delta x$ and $\Delta y$ cartesian control actions (first two dimensions of the action vector) for our analysis.
Let $A \in \mathbb{R}^{200 \times 2}$ denote the inferred actions and $B \in \mathbb{R}^{200 \times 2}$ denote the ground truth actions.
Based on this, we can solve a linear least squares problem to identify the linear transformation $W^\star \in \mathbb{R}^{2 \times 2}$ that maps inferred actions $A$ to real actions $B$,
\[
    W^\star = \underset{W \in \mathbb{R}^{2 \times 2}}{\text{argmin}}\ \lVert A W - B \rVert_2.
\]
The mean absolute prediction error for all camera position is roughly $1.6$cm (compared to a ground truth delta pose of roughly $5$cm), suggesting that the error is systematic.
In addition, we observe that for each camera position, the matrix $W^\star$ has condition number $\approx 1.5$, i.e., modulo a fixed scalar coefficient, $W^\star$ is approximately a rotation matrix, and thus we can compute the rotation error in the inferred coordinate axis by using
\[
    W^\star \approx \overline{W}^\star =
    \begin{bmatrix}
        \cos \theta & -\sin \theta \\
        \sin \theta & \cos \theta
    \end{bmatrix},
\]
with $\overline{W}^\star \coloneq U V^\top$ where $U$ and $V$ are the left and right singular vectors of $W^\star$, respectively.

\Cref{fig:wm_calibration} shows the camera position plotted against the rotation error in the \acModel inferred coordinate axis.
We observe that the rotation error in the inferred coordinate axis is almost a linear function of the camera position.
We can most clearly see the effects of rotation errors in the inferred coordinate axis in our single-goal reaching experiments in \Cref{fig:robot-reach}.
While the model is always able to move the arm within $4$ cm of the goal based on visual feedback from the monocular RGB camera, rotation errors in the inferred coordinate axis result in relatively suboptimal actions at each planning step, yielding a non-maximal, albeit monotonic, decrease in the distance to goal at each step.

Interestingly, since errors in the inferred coordinate axis are primarily rotation-based, one can use this approach to ``calibrate'' their world model by simply rotating all inferred actions by $W^\star$, and thereby introduce the desired invariance to camera position.
Such an unsupervised calibration phase would involve the robot performing random actions, solving a linear least squares problem by comparing its inferred optimal actions to the actual actions it executed, and then multiplying its inferred actions by the rotation matrix before sending them to the controller during task execution.
While such an approach is interesting, we emphasize that we \emph{do no such calibration} in our experiments.

\section{Visual Classification} \label{app:probe_understanding}

We describe in more detail the evaluation procedure used for the classification tasks described in \Cref{sec:encoder_comparison}.

\subsection{Hyperparameters} \label{app:probe_understanding_hp}

\paragraph{\bf Probe Architecture.}
We train an attentive probe on top of the frozen encoder output using the training data from each downstream task. Our attentive probe is composed of four transformer blocks, each using 16 heads in the attention layer. The first three blocks use standard self-attention; the final block uses a cross-attention layer with a learnable query token. The output of the cross-attention layer in the final block is added back to the query token as a residual connection before applying the rest of the block (LayerNorm, followed by MLP with a single GeLU activation). The transformer blocks are followed by a final linear classifier layer.

\paragraph{\bf Evaluation setup parameters.}

All models follow the same evaluation protocol and use a resolution of $256\times256$, except our \model ViT-g$_{384}$. For video evaluations, we sampled multiple clip segments from each input video. During validation, we also extracted three spatial views from each segment (instead of one view during training). The number of clip segments, frame step parameter, and global batch size vary for each eval; parameters used for each evaluation can be found in \Cref{tb:understanding_params}. By default, we use $16\times2\times3$ inputs for SSv2 (16 frames clip, 2 temporal crops, 3 spatial crops), $16\times8\times3$ for K400, $32\times8\times3$ for COIN, and $32\times4\times3$ for Diving-48 and Jester. \model ViT-g$_{384}$ uses a higher resolution of $384\times384$ for K400, COIN, Diving-48, and Jester, $512\times512$ for ImageNet and $384\times384$ with $64\times2\times3$ inputs for SSv2.

\paragraph{\bf ImageNet evaluation.} For ImageNet, we repeat each input image to produce a 16-frame video clip. We also use a larger global batch size (1024 instead of 256 or 128), and do not use multiple clips or views per sample.

\paragraph{\bf Jester and Diving-48 evaluation.} Our Jester and Diving-48 action classification evaluation tasks differ from the other understanding evaluations in several ways, primarily in that we employ a multilayer strategy. Instead of attending to the tokens from only the last layer of the encoder, we extract tokens from four encoder layers (the last layer and three intermediate layers) and attend to all of them. (\Cref{tb:understanding_layers} shows the layers we used for each encoder size.) We also train the probes for these two evaluations with only three classification heads (instead of 20 for the other evaluations), but train for 100 epochs (instead of 20) as these evaluations benefit from longer training. We use a global batch size of 128 for both evaluations.

\paragraph{\bf Optimization.} For each evaluation, we simultaneously train multiple classifier heads with different hyperparameters (learning rate and weight decay), reporting the accuracy of the best-performing classifier. For most of our evaluations (Kinetics, SSv2, COIN, and ImageNet), we train for 20 epochs and use 20 heads, each using one of five learning rate values and four weight decay values, and the learning rate decays according to a cosine schedule. We provide a summary of all hyperparameters in \Cref{tb:understanding_params}.

\begin{table}[t]
    \small
    \caption{\small{\bf Visual Classification eval params.} Default parameters used for the visual classification evaluations, with non-default values for each eval (* denotes default). All attentive probes use 4 transformer blocks with 16 heads.}
    \centering
    \label{tb:understanding_params}
    \begin{tabular}{l c c c c c}
      \bf Parameter & \bf Default (K400) & \bf ImageNet & \bf SSv2 & \bf COIN & \bf Jester/Diving-48 \\
        \toprule
        Number of frames & 16 & 16 & 16 & 32 & 32 \\
        Segments / Clip & 8 & 1 & 2 & 8 & 4 \\
        Views / Segment & 3 & 1 & * & * & * \\
        Frame Step & 4 & n/a & * & *  & 2 \\
        Epochs & 20 & * & * & * & 100 \\
        Batch Size (global) & 256 & 1024 & * & 128 & 128 \\
        Resolution & $256\times256$ & * & * & * & * \\
        Classifier Heads & 20 (4x5) & * & * & * & 3 (3x1) \\
        Classifier Learning Rates & [5e-3 3e-3 1e-3 3e-4 1e-4] & * & * & * & [1e-3 3e-4 1e-4] \\
        Classifier Weight Decay & [.8 .4 .1 .01] & * & * & * & [.8] \\
        \bottomrule
    \end{tabular}
\end{table}
\begin{table}[t]
    \small
    \caption{\small{\bf Input layers for Jester/Diving-48.} \small For each encoder size, indices of the four encoder layers whose tokens are used as input to the linear classifier in the Jester and Diving-48 evaluations.}
    \centering
    \label{tb:understanding_layers}
    \begin{tabular}{l c c}
        \bf Encoder & \bf \# Layers & \bf Attended Layers \\
        \toprule
        \bf ViT-L & 24 & 17, 19, 21, 23 \\
        \bf ViT-H & 32 & 25, 27, 29, 31 \\
        \bf ViT-g & 40 & 24, 29, 34, 39 \\
        \bottomrule
    \end{tabular}
\end{table}

\subsection{Additional Results} \label{app:probe_understanding_add_res}

\paragraph{\bf Probe Size.} Since we use a four-layer attentive probe for these evaluations, we investigate whether using a smaller probe impacts evaluation performance. We re-run our six understanding evaluations (for two model sizes, ViT-L and ViT-g) with a smaller probe consisting of a single cross-attention block using 16 attention heads. Unlike in~\Cref{sec:encoder_comparison}, we use 16 frames for all evaluations, including Diving-48 and Jester. See \Cref{tb:probe_size} for classification performance---we confirm that our four-layer probe outperforms a single-layer attentive probe across all understanding evaluations (except for Jester), by an average of $+1.4$ points accuracy for ViT-L and $+1.0$ points for ViT-g.

\paragraph{\bf Impact of encoder multilayer.} We study the impact of feeding tokens from multiple layers from the encoder to the attentive probe during evaluation. \Cref{tab:encoder_multilayer} shows that Diving-48 and Jester strongly benefit from information from deeper layers of the encoder.

\begin{table}[t]
    \small
    \caption{\small{\bf Encoder Multilayer Ablation.} \small We vary the number of encoder layers fed to the attentive probe. We report the classification performances of attentive probes trained on top of \model with 16 frames at $256\times256$ resolution.}
    \centering
    \label{tab:encoder_multilayer}
    \begin{tabular}{l c c c}
        \\
        \toprule
        \bf Model & \bf Encoder Layers & \bf Diving-48  & \bf Jester \\
        \toprule
        \bf ViT-g & 1 & 82.9 & 96.1 \\
        \bf ViT-g & 4 & 86.7 & 97.6 \\
        \bottomrule
    \end{tabular}
\end{table}

\begin{table}[t]
    \small
    \caption{\small{\bf Probe Size Ablation.} \small We vary the number of layers in the attentive probe. We report the classification performances of attentive probes trained on top of \model with 16 frames at $256\times256$ resolution.}
    \centering
    \label{tb:probe_size}
    \begin{tabular}{l c c c c c c c c}
        \\
        \toprule
        & & & \multicolumn{3}{c}{\bf\it Motion Understanding} &  \multicolumn{3}{c}{\bf\it Appearance Understanding}\\
        \bf Model & \bf Probe Layers & \bf Avg. & \bf SSv2  & \bf Diving-48  & \bf Jester &  \bf K400 & \bf COIN & \bf IN1K\\
        \toprule
        \bf ViT-L & 1 & 84.0 & 72.0 & 83.2 & 97.7 & 83.3 & 85.9 & 81.8\\
        \bf ViT-L & 4 & 85.6 & 73.6 & 87.1 & 97.7 & 85.1 & 86.8 & 83.5\\
        \bf ViT-g & 1 & 86.0 & 74.8 & 85.3 & 97.8 & 85.6 & 88.9 & 83.5\\
        \bf ViT-g & 4 & 87.0 & 75.6 & 86.7 & 97.6 & 86.6 & 90.7 & 84.6\\
        \bottomrule
    \end{tabular}
\end{table}

\section{Action Anticipation} \label{app:action_anticipation}

We provide additional details, results, and ablations related to the Epic-Kitchen 100 action anticipation evaluation of \Cref{section:prediction}.

\subsection{Hyperparameters} \label{app:action_anticipation_hp}

\paragraph{\bf Probe Architecture.} Our probe architecture for action anticipation follows the architecture of our classification probe described in \Cref{app:probe_understanding_hp}, consisting of four transformer blocks, including a last cross-attention layer with a set of learnable query tokens, followed by a final linear classifier layer for each query token.

\paragraph{\bf Evaluation setup parameters.} 
We use a focal loss~\citep{lin2017focal} with a $\alpha=0.25$ and $\gamma=2.0$ when training the probe; this loss is more suited for training with long-tailed imbalanced class distributions.
We use a context of 32 frames with a frame-rate of 8 frames per second at resolution $256\times256$ for \model ViT-L, ViT-H, and ViT-g; and resolution $384\times384$ for \model ViT-g$_{384}$.
During probe training, we randomly sample an anticipation time between $0.25$ and $1.75$ seconds, and an anticipation point between $0.0$ and $0.25$.
The \emph{anticipation point} identifies a point in the action segment from which to perform anticipation; i.e., an anticipation point of $0$ means that we predict the representation of the first frame in the action segment using our \model predictor before feeding it to the probe, whereas an anticipation point of $1$ means that we predict the representation of the last frame in the action segment before feeding it to the probe.
The validation anticipation time is set to 1 second and the validation anticipation point is set to $0$.
We provide a summary of the hyperparameters, including the optimization parameters in \Cref{tab:action_anticipation_params}.

\begin{table}[ht!]
    \small
    \caption{\small{\bf Action Anticipation Evaluation params.} Default parameters used for the EK100 Action Anticipation evaluation.}
    \centering
    \label{tab:action_anticipation_params}
    \begin{tabular}{l c c c c c}
      \bf Parameter & \bf EK100 \\
        \toprule
        Train Anticipation time & 0.25s -- 1.75s \\
        Train Anticipation point & 0.0 -- 0.25 \\
        Val Anticipation time & 1s \\
        Val Anticipation point & 0.0 \\
        Number of frames & 32 \\
        Frames per second & 8 \\
        Epochs & 20 \\
        Warmup epochs & 0 \\
        Batch Size (global) & 128 \\
        Classifier Heads & 20 (4x5)  \\
        Classifier Learning Rates & [5e-3 3e-3 1e-3 3e-4 1e-4] \\
        Classifier Weight Decay & [1e-4 1e-3 1e-2 1e-1]  \\
        \bottomrule
    \end{tabular}
\end{table}

\subsection{Additional results} \label{app:action_anticipation_results}

\paragraph{\bf Impact of Architecture.} \Cref{tb:ek100_arch} investigates the impact of providing the output of the \model encoder, predictor, or both, to the action anticipation probe.  Using encoder outputs already leads to competitive performance on the EK100 task. Adding the predictor provides a small but consistent improvement across action, verb, and object categories. In addition, using predictor outputs yields a non-trivial performance, but still at a much lower point compared to using the encoder, showing that the EK100 task mostly requires strong semantic understanding, as opposed to forecasting capabilities.

\begin{table}[t]
  \centering
    \small
    \caption{\small{\bf EK100: Impact of Anticipation Probe Inputs.} We investigate the impact of providing the outputs of the \model encoder, predictor, or both, to the action anticipation probe.  Using encoder outputs already leads to competitive performance on the EK100 task. Adding the predictor provides a small but consistent improvement across action, verb and object categories.}
    \label{tb:ek100_arch}
    \begin{tabular}{c c c c c c}
    & & \multicolumn{3}{c}{\bf\it Action Anticipation} \\
    \bf Encoder & \bf Predictor & \small Verb & \small Noun & \small Action \\
    \toprule
    \checkmark &                & 61.3 & 57.0 & 39.1 \\
    & \checkmark                & 48.7 & 34.7 & 20.2 \\
    \checkmark & \checkmark     & 63.6 & 57.1 & 39.7 \\
\end{tabular}
\end{table}

\begin{figure}[ht!]
    \centering
    \small
    \begin{subfigure}[b]{0.32\textwidth}
        \centering
        \includegraphics[width=\linewidth]{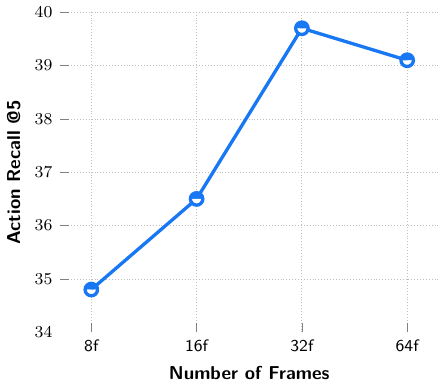}
    \end{subfigure}\hfill
    \begin{subfigure}[b]{0.32\textwidth}
        \centering
        \includegraphics[width=\linewidth]{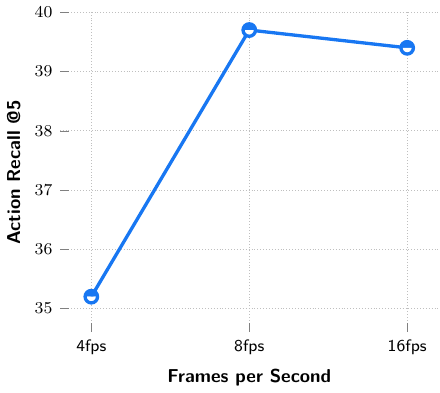}
    \end{subfigure}\hfill
    \begin{subfigure}[b]{0.32\textwidth}
        \centering
        \includegraphics[width=\linewidth]{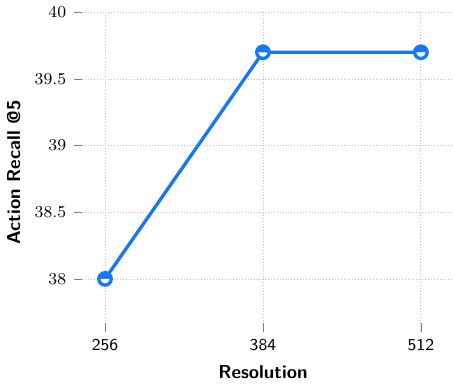}
    \end{subfigure}
    \caption{
    {\bf Protocol ablation for action anticipation on EK100.}
    {\bf (Left)} Performance with respect to the number of context frames used for action anticipation.
    {\bf (Middle)} Performance with respect to the frame rate (fps) used for inference; number of context frames fixed at 32.
    {\bf (Right)} Performance with respect to the spatial resolution (height and width) of the context frames used for action anticipation.}
    \label{fig:ek100_resolution}
\end{figure}

\begin{figure}[ht!]
    \centering
    \begin{subfigure}[b]{0.48\textwidth}
        \centering
        \includegraphics[width=\linewidth]{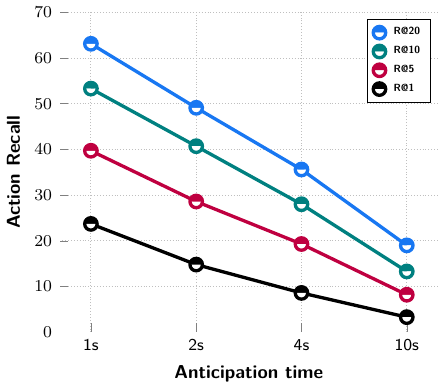}
    \end{subfigure}\hfill
    \begin{subfigure}[b]{0.48\textwidth}
        \centering
        \includegraphics[width=\linewidth]{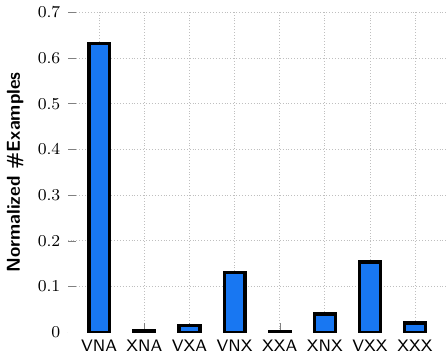}
    \end{subfigure}
    \caption{\small{\bf (Left): Impact of longer-term anticipation times.} Performance on EK100 action anticipation, at several recall values and anticipation times. {\bf (Right): Distribution of success and failure cases of \model}. Calculated on the validation set of EK100. VNA means that verb, noun and action are all correctly classified by the model. An X symbol means that the corresponding attribute is not correctly classified by the model. Note: the probe is composed of 3 independent classifiers for verb, noun and action, hence why the model can have a different prediction for the action and for the verb/noun pair.}
    \label{fig:ek100_longerterm_failures}
\end{figure}

\paragraph{\bf Impact of Input Resolution.} We report in \Cref{fig:ek100_resolution}, the impact of input resolution and frame sampling parameters. In summary, \model benefits from a longer context, a higher frame rate, and higher resolution, up to a point where the performance saturates or slightly decreases. The optimal performance is obtained by training with a 32-frames context length, a frame rate of 8 and a resolution of $384\times384$.

\paragraph{\bf Longer-term Prediction.} We report in \Cref{fig:ek100_longerterm_failures} (Left), the impact of predicting at a longer horizon, by varying the anticipation time between (1s, 2s, 4s, 10s). For each anticipation time, we report recall at several values (1, 5, 10, 20). The results show that the performance sharply decreases as the anticipation time increases, which is expected since forecasting the future in EK100 is a non-deterministic task. 

\paragraph{\bf Analysis of Failure Cases.} We report in \Cref{fig:ek100_longerterm_failures} (Right), the distribution of failure and success prediction, on the EK100 validation set, between each configuration of success/failure for verb, noun, and action. The model performs very well, and the most represented configuration is, therefore, a full success across verb noun and action. The most represented failure configurations all include a failure to find the action.

\section{Video Question Answering}
\label{app:language_alignment}

In this section, we provide details on training a \model Multi-modal Large Language Model (MLLM). We follow the LLaVA framework \citep{liu2024improved} to train the MLLM, where the vision backbone is using \model, and the LLM backbone can be any off-the-shelf pretrained LLM, akin to the \textit{non-tokenized early fusion} \citep{wadekar2024evolution} setup.  The MLLM ingests the output embeddings of the vision encoder, which are projected to the hidden dimension of the LLM backbone using a \textit{projector} module. The \textit{projector}  is typically a 2-layer MLP.  The MLLM  is trained using a mix of image-text and video-text paired data, in a series of progressive training steps.

To understand the impact of data scale, we use a dataset of 88.5 million image-text and video-text pairs, similar to what was used for training PerceptionLM~\citep{cho2025perceptionlm}. As mentioned in \Cref{section:language_understanding}, we investigate two setups: (a) \textit{controlled}, where we train with 18M image and video-text pairs, and we evaluate \model and other encoders on the exact same MLLM training setup, and (b) \textit{scaling}, where we take \model VITg$_{384}$ and use the full aligment dataset.
To further test the versatility of \model, we use Qwen2-7B-Instruct \citep{yang2024qwen2technicalreport} as the language backbone for the controlled experiments, and Llama 3.1 8B Instruct \citep{grattafiori2024llama} for the scaling experiments. We describe the training details in the following sections.

\subsection{Processing Images and Videos as Input}
Since video question answering uses video instead of image inputs, the number of output visual tokens increases significantly compared to image question answering. If required, we can use pooling methods to reduce the number of visual tokens. Popular pooling methods involve adaptive 2x2 pooling \citep{cho2025perceptionlm}, Perceiver Sampler \citep{jaegle2021perceiver}, Attentive Pooling \citep{bardes2024revisiting}, etc. 

Additionally, we observed that learning from image-text pairs is crucial for high performance in downstream benchmarks. In order to train with images, a simple approach is to repeat the given image for $k$ frames, where $k$ is the maximum amount of frames supported by \model. However, during our initial experiments we find this strategy is ineffective at improving downstream performance, as it does not allow the model to extract fine-grained information.
Therefore, we employ a modified \textit{Dynamic $S^2$} strategy introduced by \citet{liu2024nvila} to provide \model higher resolution granularity during training. This method adaptively processes an image at native resolution with different aspect ratios to preserve their  original resolution, by creating a sequence of tiles of maximum size supported by \model.
In case of videos, we choose to train with a fixed number of frames $f_n$, by balancing the number of visual tokens with compute budget.

\subsection{Controlled Setup}

\paragraph{\bf Training details} 
For the controlled setup, we follow the LLaVA-NEXT framework \citep{liu2024improved, zhang2024llavanext-video}, where we use Qwen2-7B-Instruct \citep{yang2024qwen2technicalreport} as the base LLM  for all encoders.
To reduce the number of visual tokens, we employ an attentive pooler with a factor of 4-16, depending on the compute budget and the number of visual patches. See \autoref{tb:llm_training_details} for more details.

Our training setup follows the LLaVA-NeXT pipeline \citep{li2024llava}, which consists of multiple staged training phases. Concretely, the stages consist of: a) aligning the attentive pooler with image captioning data (Stage 1), b) training the full model on high quality image captioning (Stage 1.5), and c) training the full model on large scale image question answering (Stage 2). We add an extra stage to train on large scale video captioning and question answering (Stage 3). 
We use 18 million image and video-text aligned data. 
The LLM progressively improves its understanding of the visual tokens after multiple staged training, with the biggest improvement in video question answering tasks after Stage 3. 

We explore \emph{frozen} and \emph{finetuned} encoder alignment setups.
In both setups, full parameters of the LLM and projector are trained, and in the latter, the \model parameters are additionally unfrozen. 
To reduce the number of visual tokens and keep the MLLM context length fixed, we employ an attentive pooler as the projector to reduce the number of visual tokens by a factor of 4, unless otherwise denoted.
The implementation used for this controlled study is based on the Llava-NEXT codebase,\footnote{\href{https://github.com/LLaVA-VL/LLaVA-NeXT}{https://github.com/LLaVA-VL/LLaVA-NeXT}} and uses Pytorch 2.5.1, Transformers 4.46.0, Flash attention 2 and DeepSpeed 0.14.4 for model implementation, faster training and multi-gpu model sharding respectively. We train all models using 128 H100 GPUs with an effective batch size of 256 across all stages. We perform all optimizations using AdamW with 0 weight decay. For Stages 1 and 1.5, we use learning rate of 1e-5 with cosine decay, and for Stages 2 and 3 we use constant learning rate of 5e-6. In all stages, we use linear warmup for the first 3\% of training steps.
Training hyperparameters are listed in \autoref{tb:llm_training_details}.

\paragraph{\bf Baselines.} To assess the ability of \model to capture spatiotemporal details for VidQA, we compare to leading off-shelf image encoders. Specifically, we compare to DINOv2~\citep{oquab2023dinov2}, SigLIP2~\citep{tschannen2025siglip}, and Perception Encoder~\citep{bolya_perception_encoder_2025}. DINOv2 is a self-supervised image model, while SigLIP2 and Perception Encoder are both trained with language supervision using noisy image-text captions. We apply all image encoders at their ``native'' pretrained resolution, which is 518px, 384px, and 448px, respectively, on each video frame independently.

We keep all training details the same, except that we increase the attentive pooling ratio to 16 to keep the number of image tokens relatively similar among models. See \autoref{tb:llm_training_details} for details.

\begin{table}[]
\centering
\small
\caption{\small{\bf Hyperparameters for controlled comparison of vision encoders. We use each vision encoder with its native pretrained input resolution.}}
\label{tb:llm_training_details}
\fontsize{8pt}{8pt}\selectfont
\begin{tabular}{l c c c c c c c}
\toprule
Model & Pooling Ratio & Vision Tokens & BS & Stage 1/1.5 LR & Stage 2/3 LR & WD & Frames \\
\midrule
\model ViT-L$_{256}$ & 4 & 4096 & 256 & \multirow{5}{*}{1e-5} & \multirow{5}{*}{5e-6} & \multirow{5}{*}{0.0} & \multirow{5}{*}{128} \\
\model ViT-H$_{256}$ & 4 & 4096 & 256 & & & & \\
\model ViT-g$_{256}$ & 4 & 4096 & 256 &  & & & \\
\model ViT-g$_{384}$ & 4 & 9216 & 256 & & & & \\
\model ViT-g$_{512}$ & 8 & 8192 & 256 & & & & \\
\midrule
DINOv2$_{518}$ & 16 & 10952 & 256 & \multirow{4}{*}{1e-5} & \multirow{3}{*}{5e-6} & \multirow{3}{*}{0.0} & \multirow{4}{*}{128} \\
SigLIP2$_{384}$ & 16 & 5832 & 256 & & & & \\
PE$_{448}$ & 16 & 8192 & 256 & & & & \\
\bottomrule
\end{tabular}
\end{table}

\begin{figure}[!h]
    \centering
    \small
    \includegraphics[width=0.4\linewidth]{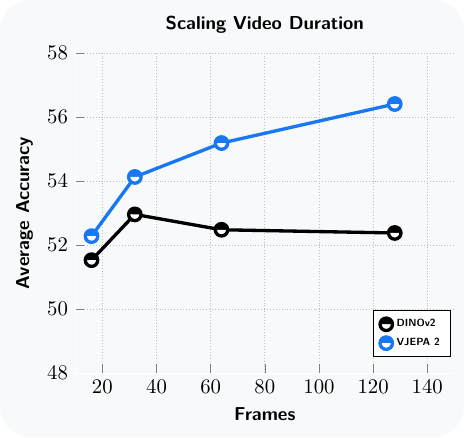}
    \caption{{\bf Impact of video duration during visual instruction tuning.}
    We investigate the effect of increasing the number of frames during visual instruction tuning, where the encoder is frozen. We observe that with more frames, \model performance linearly increases compared to DINOv2, an SSL-based image encoder, showing the potential of \model to scale with more frames.
    }
    \label{fig:llm_scaling}
\end{figure}

\paragraph{\bf Evaluation.} To evaluate the capability of \model to understand the world through video and language, we select popular evaluation datasets built to test spatio-temporal reasoning abilities. 
To ensure reproducible evaluation, we utilize the \texttt{lmms-eval} library \citep{lmms_eval2024,zhang2024lmmsevalrealitycheckevaluation} to conduct our experiments, which is a vision model enabled fork of \texttt{llm-eval-harness} \citep{eval-harness}, which is a popular evaluation library for evaluating LLMs on text-based tasks.
In the controlled setup, for each model and dataset, we evaluate by using uniform frame sampling mechanism, and choosing 128 frames during inference.
For PerceptionTest, we further train the model for 5 epochs on the training set.

\paragraph{\bf Impact of Video Duration} In the controlled setup, we perform an analysis to understand \model's capability in long-form video understanding. We train MLLMs on \model and DINOv2, keeping the encoders frozen, and by increasing the number of frames we use in training and testing. 
We observe as the number of frames increases, performance on downstream tasks \textit{linearly} improves for \model, but decreases and remains flat in case of DINOv2 (\autoref{fig:llm_scaling}).
This highlights the potential of video encoders such as \model to understand long-form videos with natural language queries, via adapting an LLM using \model as the visual encoder.

\subsection{Data scaling setup}

\begin{table}[h]
    \small
    \caption{\small{\bf Data scaling training parameters.} }
    \centering
    \label{tab:data_scaling_params}
    \begin{tabular}{l c}
      \bf Parameter & \bf Values \\
        \toprule
        \multicolumn{2}{l}{\bf\it\underline{Common parameters}}\\[1ex]
        Crop Size & 384 \\
        Video Frames per Second & 1 \\
        Sampling method & Uniform \\
        Seed & 777 \\
        \multicolumn{2}{l}{\bf\it\underline{Stage 1}}\\[1ex]
        Steps & 16000 \\
        Warmup Steps & 96 \\
        Batch Size (global) & 128 \\
        Learning Rate & 1e-4 \\
        Final Learning Rate & 1e-6 \\
        Weight Decay & 0.05 \\
        Max sequence length & 1920 \\
        \multicolumn{2}{l}{\bf\it\underline{Stage 2}}\\[1ex]
        Steps & 35000 \\
        Warmup Steps & 200 \\
        Batch Size (global) & 2048 \\
        Learning Rate & 4e-5 \\
        Final Learning Rate & 4e-7 \\
        Weight Decay & 0.05 \\
        Max sequence length & 6400 \\
        Image tiles & 16 \\
        Video frames & 16 \\
        \multicolumn{2}{l}{\bf\it\underline{Stage 3}}\\[1ex]
        Steps & 28000 \\
        Early stopping step & 22000 \\
        Warmup Steps & 168 \\
        Batch Size (global) & 2048 \\
        Learning Rate & 1e-5 \\
        Final Learning Rate & 1e-7 \\
        Weight Decay & 0.05 \\
        Max sequence length & 12800 \\
        Image tiles & 32 \\
        Video frames & 32 \\
        \bottomrule
    \end{tabular}
\end{table}

\paragraph{\bf Training details} 
In the scaling setup, we follow the framework used by \citet{cho2025perceptionlm} to train Perception LM 8B. Specifically, we utilize the released codebase, which is based on Lingua \citep{meta_lingua}. We modify the code to use \model encoder, and we use the Llama 3.1 8B Instruct \citep{grattafiori2024llama} as the backbone LLM. Unlike \citet{cho2025perceptionlm}, we do not use pooling, instead we train \model VIT-g$_{384}$ using MLP projector, leading to 288 tokens per frame. The training setup also consists of three progressive stages: Stage 1: aligning the MLP pooler with image captioning data; Stage 2: training on a mix of image-text captioning and QA data; and Stage 3) training on video-text captioning and QA data.
We scale up the data size to 88.5 million samples.
Our setup uses Pytorch 2.5.1 and Perception LM training code,\footnote{\href{https://github.com/facebookresearch/perception_models/tree/main}{https://github.com/facebookresearch/perception\_models}} modified with the \model encoder. We train on 512 H100 GPUs for Stage 2 and Stage 3 with a global batch size of 2048 and 1024 respectively. Details of the training hyperparams are provided in \autoref{tab:data_scaling_params}.

\paragraph{\bf Baselines.}
We compare our scaling runs with Qwen2VL \citep{wang2024qwen2}, Qwen2.5VL \citep{bai2025qwen2}, InternVL-2.5 \citep{chen2024expanding}, and PerceptionLM 8B \citep{cho2025perceptionlm}. Baseline numbers are sourced directly from the papers, except for MVP which we run ourselves.

\paragraph{\bf Evaluation.}
We follow similar evaluation pipeline as reported in the controlled setup, using \texttt{lmms-eval} library. 
We report our model evaluations on 32 frames.

\end{document}